\documentclass[10pt,twocolumn,letterpaper]{article}

\usepackage[pagenumbers]{iccv}

\newcommand{\cmark}{\textcolor{green}{\ding{51}}}  
\newcommand{\xmark}{\textcolor{red}{\ding{55}}} 

\usepackage{colortbl}

\usepackage[acronym]{glossaries}
\newacronym[longplural={Mixture of Experts}]{moe}{MoE}{Mixture of Experts}
\newacronym{dl}{DL}{Deep Learning}
\newacronym{nlp}{NLP}{Natural Language Processing}
\newacronym{cv}{CV}{Computer Vision}
\newacronym{vit}{ViT}{Vision Transformer}
\newacronym{deit}{DeiT}{Data-efficient image Transformer}

\newacronym{vbp}{VBP}{Variance-Based Pruning}
\newacronym{swin}{Swin}{Swin-Transformer}
\newacronym{convnext}{ConvNeXt}{ConvNeXt}
\newacronym{nvit}{NViT}{Global Vision Transformer Pruning with Hessian-Aware Saliency}
\newacronym{tome}{ToMe}{Token Merging}

\newacronym{sota}{SoTA}{state-of-the-art}

\newacronym{cnn}{CNN}{Convolutional Neural Network}
\newacronym{mhsa}{MHSA}{Multi-Head Self-Attention}
\newacronym{mlp}{MLP}{Multi-Layer Perceptron}
\newacronym[longplural={Mixture of Extracted Experts}]{moee}{MoEE}{Mixture of Extracted Experts}
\newacronym{hdbscan}{HDBSCAN}{Hierarchical Density Based Spatial Clustering for Applications with Noise}
\newacronym{pca}{PCA}{Principal Component Analysis}
\newglossaryentry{encblock}{name={encoder block}, description={encoder block}}
\newglossaryentry{imagenet}{name={ImageNet-1k}, description={ImageNet-1k}}
\newglossaryentry{cifar100}{name={CIFAR-100}, description={CIFAR-100}}
\newglossaryentry{linlayer}{name={linear layer}, description={linear layer}}
\newglossaryentry{layernorm}{name={layer normalization}, description={layer normalization}}
\newglossaryentry{actfunc}{name={activation function}, description={activation function}}

\newacronym{wlog}{w.l.o.g.}{without loss of generality}
\glsunset{convnext}

\glsaddall
\glsdisablehyper

\usepackage{amsmath}
\usepackage{amssymb}

\usepackage{pifont}

\usepackage{tabularx} 
\usepackage{booktabs}

\usepackage{graphicx}
\usepackage{float}

\usepackage{subcaption}

\usepackage{siunitx}

\usepackage{textcomp}

\usepackage[symbol]{footmisc}
\newcommand{\printfnsymbol}[1]{\textsuperscript{#1}}

\usepackage{csquotes}

\usepackage{arydshln}

\definecolor{iccvblue}{rgb}{0.21,0.49,0.74}
\usepackage[pagebackref,breaklinks,colorlinks,allcolors=iccvblue]{hyperref}

\def\paperID{14403}
\def\confName{ICCV}
\def\confYear{2025}

\title{Variance-Based Pruning for Accelerating and Compressing Trained Networks}

\author{Uranik Berisha\printfnsymbol{1}\printfnsymbol{2}, Jens Mehnert\printfnsymbol{1} and Alexandru Paul Condurache\printfnsymbol{1}\printfnsymbol{2}\\
	{\small \printfnsymbol{1}Automated Driving Research, Robert Bosch GmbH, 70469 Stuttgart, Germany }\\
	{\small \printfnsymbol{2}Institute for Signal Processing, University of L{\"u}beck, 23562 L{\"u}beck, Germany}\\
	{\tt\small \{Uranik.Berisha,JensEricMarkus.Mehnert,AlexandruPaul.Condurache\}@de.bosch.com}
}

\begin{document}
\maketitle
\begin{abstract}
	
Increasingly expensive training of large models motivate reusing the vast library of already trained state-of-the-art networks. However, their latency, high computational costs and memory demands pose significant challenges for deployment. While structured pruning methods can reduce these factors, they often require costly retraining or even training from scratch to recover the lost accuracy resulting from the structural modifications. Maintaining the provided performance of trained models after structured pruning and thereby avoiding extensive retraining remains a challenge.

To solve this, we introduce Variance-Based Pruning, a simple and structured one-shot pruning technique for efficiently compressing networks, with minimal finetuning. Our approach first gathers activation statistics, which are used to select neurons for pruning. Simultaneously the mean activations are integrated back into the model to preserve a high degree of performance. On \gls{imagenet} recognition tasks, we demonstrate that directly after pruning DeiT-Base retains over 70\% of its original performance and requires only 10 epochs of fine-tuning to regain 99\% of the original accuracy while simultaneously reducing MACs by 35\% and model size by 36\%, thus speeding up the model by 1.44$\times$. The code is available at: \texttt{\url{https://github.com/boschresearch/variance-based-pruning}}

\end{abstract}    
\begin{figure*}[t]
	\centering
	\begin{subfigure}[t]{0.32\textwidth} 
		\centering
		\includegraphics[width=\textwidth]{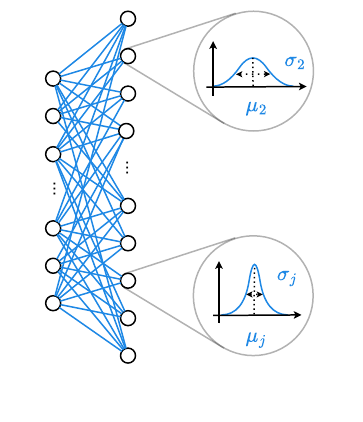}
		\caption{We compute neuron-wise mean and variance of activations at the hidden layer of each MLP.}
		\label{fig:activation-collection}
	\end{subfigure}
	\hfill
	\begin{subfigure}[t]{0.32\textwidth} 
		\centering
		\includegraphics[trim=0 0 0 0, clip, width=\textwidth]{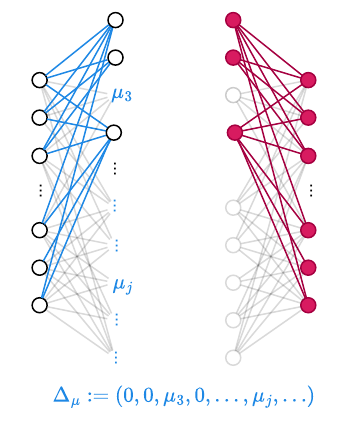}
		\caption{We prune neurons with the smallest variances and replace their activation with the mean.}
		\label{fig:variance-pruning}
	\end{subfigure}
	\hfill
	\begin{subfigure}[t]{0.32\textwidth} 
		\centering
		\includegraphics[width=\textwidth]{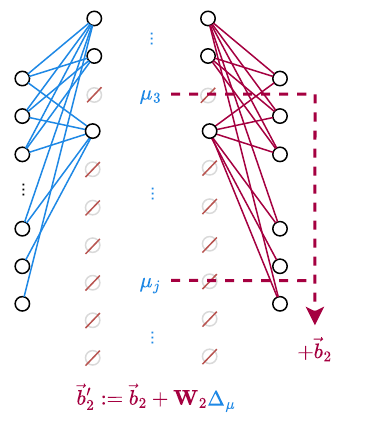}
		\caption{We shift the mean activation of the removed neurons into the bias of the next layer.}
		\label{fig:mean-shift}
	\end{subfigure}
	\caption{Schematic visualization of the steps of Variance-Based Pruning. (a) Activation Statistics Computation, (b) Variance-Based Pruning, (c) Mean-Shift Compensation}
	\label{fig:variance-based-pruning}
\end{figure*}

\section{Introduction}

In both \gls{cv} and \gls{nlp}, Transformers have become the dominant architecture~\citep{aiiw1wtfiras, tdeit&dta}. In \gls{cv}, \glspl{vit}~\citep{aiiw1wtfiras} and their derivatives, such as \glspl{deit}~\citep{tdeit&dta}, \glspl{swin}~\citep{sthvtusw} and \glspl{convnext}~\citep{acft2}, achieve remarkable results but at significant costs in three main areas: training, storage, and inference. These large models require long training times, often spanning hundreds of epochs to achieve their performance, extensive memory to store their parameters, and high computational resources for inference causing high latency~\citep{tdeit&dta}. 

Fixing all three issues simultaneously remains especially challenging.
For example, using the available collection of trained networks provided by the research community is a straightforward way to solve the first problem of costly training. 
However, this does not address the storage and inference costs, which can still make the use of these trained networks impractical for deployment on resource-constrained hardware \citep{spodcnn}.
To address these demands, various model pruning techniques have been proposed. These range from removing individual weights within layers in unstructured pruning~\citep{obd} to removing entire neurons, layers, or blocks in structured pruning~\citep{spodcnn}. 

While unstructured pruning can often retain a high degree of accuracy in trained networks, it is difficult to translate these theoretical gains into real-world speed-ups \citep{spodcnn}. Since modern hardware is optimized for dense matrix multiplications, they cannot exploit the resulting sparsity effectively and therefore cannot solve the inference costs~\citep{spodcnn, obd}. 

In contrast, structured pruning leads to straightforward reductions in both inference costs and memory, but again requires extensive retraining to recover the lost accuracy resulting from these major structural modifications. For instance, \gls{nvit}~\citep{gvtpwhas} requires 300 epochs of fine-tuning despite using trained networks, highlighting how addressing memory and inference speeds alone can still demand large additional training overhead.

On the other hand, dynamic pruning approaches, such as token pruning, avoid permanently modifying the model architecture and instead dynamically reduce the number of features processed by the network~\citep{devtwdts, tmyvbf}.
This minimizes the accuracy drop after pruning, indicating that the model preserved most important features and representation, thus providing a better starting point for subsequent fine-tuning. \gls{tome}~\citep{tmyvbf} embraces this idea by merging tokens to reuse the pruned information, thus reducing the accuracy drop even further and allowing off-the-shelf deployment without any fine-tuning. While this solves the inference costs as well as the training costs, since the methods do not modify the model structure itself, the memory footprint remains unchanged \citep{tmyvbf}. This again is a limitation for hardware-constrained settings~\citep{spodcnn}. Fully resolving all three problems of training costs, storage, and inference speed simultaneously remains a challange.

In this work, we fill this gap by introducing \textbf{\gls{vbp}}, a simple and structured pruning approach that provides speed-ups and memory savings with minimal fine-tuning. In order to keep the method simple and allow deployment across a wide variety of architectures, we prune only the heavy \gls{mlp} layers within the transformer blocks. In the first step, our approach leverages activation statistics gathered in the hidden layers of the \gls{mlp} using Welford's algorithm~\citep{noamfccsosap} (see \cref{fig:activation-collection}). In the second step, these statistics are used to identify and remove the least impactful hidden neurons (see \cref{fig:variance-pruning}). In the last step, which we refer to as \textbf{Mean-Shift Compensation}, we use the mean activation values of pruned neurons to mitigate accuracy drop by redistributing these mean contributions into the bias of the output layer (see \cref{fig:mean-shift}).
This retains high levels of model performance and requires only minimal fine-tuning to then regain most of the original accuracy. 
\\
\\
\textbf{Our contributions are summarized as follows:}
\begin{itemize}
	\item We introduce \gls{vbp}, a simple low-cost structured pruning method broadly applicable across various architectures to significantly reduce computational costs while maintaining performance.
	\item We apply our method to multiple architectures, including \glspl{vit}, Swin-Transformers, and \glspl{convnext}, demonstrating its generalizability.
	\item We compare VBP to similarly straightforward neuron-importance measures adapted for structured pruning, as well as to \gls{sota} pruning approaches. We benchmark against \gls{nvit} pruning, highlighting the benefits of our accuracy retention in constrained fine-tuning settings. We also apply \gls{vbp} on top of \gls{tome} demonstrating orthogonality and achieving $2\times$ speed-ups using our hybrid approach. 
	\item We perform an extensive ablation study, demonstrating the efficacy of each component of our method and provide a sensitivity analysis for the variance as a pruning criterion.
\end{itemize}

\section{RELATED WORK}
\label{sec:related-work}

\subsection{Transformer Architectures}
Initially introduced in \gls{nlp} by \citep{aiayn}, Transformers have largely replaced \gls{cnn}, since their adaption to \gls{cv} by \citep{aiiw1wtfiras} with the \gls{vit}. \glspl{vit} processes images as a sequence of non-overlapping patches, and have since been further improved. \gls{deit} \citep{tdeit&dta} refined the training procedure to decrease the training costs and reduce the data reliance. More recent architectures \citep{ctagvtbwcsw, mvt, lavticcffi}, such as \gls{swin} \citep{sthvtusw} and the transformer-inspired \gls{convnext} \citep{acft2}, which use hierarchical designs, have built upon this architecture to improve efficiency and scalability. However, despite these advances, these large models remain computationally demanding, in part due to their dense \gls{mlp} blocks, which significantly impacts inference speed and memory requirements \citep{aiayn, svt}. 

\subsection{Model Compression}
To reduce the costs of large models, model compression efforts focus on scaling down different components of a network. Traditionally pruning aims to identify and remove redundant structures within the network \citep{obd, ipuafrtitosc}. For example, magnitude pruning eliminates weights with the smallest absolute values, while SNIP and related methods \citep{sssnpbocs, ffpbrsc} identify crucial connections using gradient-based sensitivities.
Structured pruning approaches, remove entire neurons, layers or filters producing smaller matrices that allow direct computational savings in current hardware~\citep{spodcnn, vtp, w&dpfvt, csivtaetee, gvtpwhas, xpepfvt}. 
Most closely related to our method are works exploring variance-based criteria for structured pruning, such as Molchanov et al. \citep{vdsdnn}, which performs dropout at the neuron level and can introduce sparsity in neurons with high dropout rates. Weinstein et al. \citep{vpplmvtnv} use these statistics in the initial training phase to remove neurons during training. These approaches operate on models before training, improving training time but still requiring the models to be trained from scratch.
Contrary to that, NViT \citep{gvtpwhas} utilizes trained \glspl{vit}. They apply iterative structured dimensionality reductions across the entire \gls{vit} architecture and significantly reduce costs. However, these structured modifications still have a substantial impact on performance, requiring high pruning iterations of 50 epochs and retraining costs of 300 epochs to regain accuracy.

An alternative to structurally modifying the network is dynamic pruning, which adaptively selects and processes only a subset of features or combines features during inference to reduce costs. For \glspl{cnn}, these have shown promising speed-accuracy trade-offs~\citep{cmoda, dfdcocfte}.
Analogously for transformer-like models, token pruning reduce the number of tokens to be processed~\citep{devtwdts, moddacitblm, aavtfeir, tmyvbf}. For instance, Bolya et al. \citep{tmyvbf} merges similar tokens rather than pruning them, thereby conserving most of the information and retaining high accuracy before the fine-tuning. However, by design these methods cannot reduce the memory footprint. 

\subsection{Our Approach}
Our work builds upon these ideas by combining (i) the advantages of structured pruning by removing low-variance activation neurons with (ii) the advantages of dynamic pruning by incoorperating their mean contribution into the bias of the last MLP layer.
Our approach is based on the observation that neurons within \gls{mlp} layers exhibit varying levels of contribution to the overall representation, as established in works on emerging modularity in trained Transformers for both \gls{nlp} and \gls{cv} \citep{emiptt, eddmoeeftn}. By applying our variance-based pruning approach to the widespread \gls{mlp} layers, we provide a structured method applicable across a broad range of models. Our method is particularly beneficial for transformer-like architectures, where \gls{mlp} layers heavily contribute to the computational overhead \citep{aiayn, svt}.

\section{Methodology}
\label{sec:methodology}

In this section, we detail the components involved in our proposed \gls{vbp} approach and provide a mathematical derivation, justifying our pruning criterion and compensation strategy. 
\\
Our method consists of three major steps:
\begin{enumerate}
	\item \textbf{Activation Statistics Computation} The neuron activation statistics are computed using Welford's algorithm.
	\item \textbf{Variance-Based Pruning} The neurons to be pruned are selected based on the lowest activation variance. 
	\item \textbf{Mean-Shift Compensation} The mean activations of the selected neurons are added back into the output bias.
\end{enumerate}

\subsection{Step 1: Activation Statistics Computation}
\label{sec:activation-stats}

As we aim for a simple and widely adoptable method, we prune only the \glspl{mlp}. This requires reducing the hidden layers of the network while keeping the input and output layers intact to ensure seamless integration with the rest of the model structure by maintaining dimensional consistency in the input and output.

\Glsentryshort{wlog} we therefore consider a \gls{mlp} with a single hidden layer of dimension 
\(
D_\text{hid}
\),
that maps an input vector 
\(
\mathbf{x} \in \mathbb{R}^{D_\text{in}}
\)
to an output vector
\(
\mathbf{y} \in \mathbb{R}^{D_\text{out}}
\).
Using a pointwise nonlinear activation function
\(\sigma(\cdot)\), this \gls{mlp} can then be described as a function
\begin{equation}
\label{eq:mlp-structure}
\mathbf{h} \;=\; \sigma\!\bigl(\mathbf{W}_1 \,\mathbf{x} \;+\; \mathbf{b}_1\bigr),
\quad
\mathbf{y} \;=\; \mathbf{W}_2 \,\mathbf{h} \;+\; \mathbf{b}_2,
\end{equation}
where 
\(
\mathbf{W}_1 \in \mathbb{R}^{D_\text{hid} \times D_\text{in}}, \,
\mathbf{b}_1 \in \mathbb{R}^{D_\text{hid}}, \,
\mathbf{W}_2 \in \mathbb{R}^{D_\text{out} \times D_\text{hid}}, \,
\mathbf{b}_2 \in \mathbb{R}^{D_\text{out}}
\). 
Since we deal with trained networks, these parameters \(\mathbf{W}_1,\mathbf{b}_1,\mathbf{W}_2,\mathbf{b}_2\) remain fixed during pruning, which will become important in the Mean-Shift Compensation in \cref{sec:mean-shift}.

For the hidden layer, we now want to calculate the per-neuron mean and variance vectors
\[
\boldsymbol{\mu} \;=\; \bigl(\mu_1,\dots,\mu_{D_\text{hid}}\bigr)^\top \quad\text{and}\quad
\boldsymbol{\sigma}^2 \;=\; \bigl(\sigma_1^2,\dots,\sigma_{D_\text{hid}}^2\bigr)^\top
\]

To compute these statistics, we capture the activations \(\mathbf{h}\) for every batch and use Welford's algorithm to compute a running mean \(\boldsymbol{\mu}^{(j)}\) and running second moment \(\mathbf{m}_2^{(j)}\) for an efficient online calculation of streaming data\footnote{Our method does not require a minimum number of data samples for calculating activation statistics, but using an efficient online computation allows the use of as many data samples for pruning as available.}.
Given a total of \(N\) data samples, let the \(j\)-th sample \(\mathbf{h}^{(j)} \in \mathbb{R}^{D_\text{hid}}\) be the activation vector after the nonlinearity.
Welford's algorithm ensures numerically stable calculations by updating \(\boldsymbol{\mu}^{(j)}\) and \(\mathbf{m}_2^{(j)}\) after observing the \(j\)-th sample as follows:
\begin{equation}
\label{eq:welford-1}
\boldsymbol{\mu}^{(j)} \;=\; 
\frac{j-1}{j}\;\boldsymbol{\mu}^{(j-1)} \;+\; \frac{1}{j}\;\mathbf{h}^{(j)}
\end{equation}
\begin{equation}
\label{eq:welford-2}
\mathbf{m}_2^{(j)} \;=\; \mathbf{m}_2^{(j-1)} \;+\; (\mathbf{h}^{(j)} - \boldsymbol{\mu}^{(j-1)}) \;\odot\; (\mathbf{h}^{(j)} - \boldsymbol{\mu}^{(j)})
\end{equation}
where \(\odot\) denotes elementwise multiplication. 
Once all \(N\) samples are processed, the variance is given by:
\begin{equation}
\label{eq:welford-var}
\boldsymbol{\sigma}^2 \;=\; \frac{\mathbf{m}_2^{(N)}}{N - 1}.
\end{equation}

\subsection{Step 2: Variance-Based Pruning}
\label{sec:variance-pruning}
Using this variance vector \(\boldsymbol{\sigma}^2 \in \mathbb{R}^{D_\text{hid}}\), we rank neurons by their variance values, \(\sigma_i^2\). Intuitively, if an activation rarely deviates from its mean \(\mu_i\), the corresponding neuron contributes less to the expressiveness of the network.
The pruning decision is formed across all hidden neurons in the network, to capture the neuron with the least variance across all layers.

Formally, for each \gls{mlp} \(l\), let \(D^{(l)}_{\text{hid}}\) denote the number of hidden neurons. For each neuron \(i\) in \gls{mlp} \(l\), we then simply select its activation variance  \(\sigma_{l,i}^2\) as the corresponding pruning score. Given a pruning ratio \(p\in(0,1)\), we then only need to form the global set of scores
\[
S = \bigcup_{l \in \mathcal{L}} \{ \sigma_{l,i}^2 \mid i = 1, \dots, D^{(l)}_{\text{hid}} \}.
\]
from which we select the \(p \%\) neurons with the smallest scores for pruning.

\paragraph{Optimality from a Mean-Replacement Perspective}
\label{sec:optimality}

To retain the contribution of the pruned network, we approximate the activation of a pruned neuron \( h_i \) by replacing it with its mean \( \mu_i \).
This introduces an error proportional to \( |h_i - \mu_i| \), which, by definition of the sample variance for any distribution, has an expected value:

\[
\mathbb{E}[(h_i - \mathbb{E}[h_i])^2] = \mathbb{E}[(h_i - \mu_i)^2] = \sigma_i^2.
\]

Thus, pruning the neurons with the lowest variance results in the least reconstruction error, making variance the optimal metric in this context.
This is intuitively confirmed by the fact that when the variance is zero, all activations are exactly equal to the mean.

\subsection{Step 3: Mean-Shift Compensation}
\label{sec:mean-shift}

The intuitive approach of removing neurons and replacing their activations with \(\mu_i\)\footnote{We omit the \gls{mlp} indices for clarity.} at inference time already reduces the output dimension of $\mathbf{W}_1$, but the reconstruction of the original embedding dimension \(D_{hid}\) through replacement of the pruned activations still requires carrying out the full matrix multiplication for $\mathbf{W}_2$. 

Instead, we perform an equivalent transformation by shifting these mean values directly into the bias of the final \gls{linlayer} $\mathbf{b}_2$, allowing us to reduce both the output dimension of $\mathbf{W}_1$ as well as the input dimension of $\mathbf{W}_2$, doubling the cost savings.

We start from the usual two-layer linear mapping in the \gls{mlp} block:
\[
\mathbf{y} = \mathbf{W}_2\,\mathbf{h} + \mathbf{b}_2,
\]
where \(\mathbf{h}\) is the hidden activation after pruning. For pruned neurons, we replace \(h_j\) with its mean \(\mu_j\), which we collect into a vector \(\Delta_{\mu} \in \mathbb{R}^{D_\text{hid}}\)
\[
(\Delta_{\mu})_j = 
\begin{cases}
\mu_j, & j\in\mathcal{P},\\
0, & j\notin\mathcal{P},
\end{cases}
\]
where \(\mathcal{P}\) is the set of pruned neuron indices.

Instead of explicitly inserting these means into \(\mathbf{h}\) and then multiplying by \(\mathbf{W}_2\), we exploit the linearity of the mapping:
\[
\mathbf{W}_2\,\mathbf{h}
=
\mathbf{W}_2\bigl(\mathbf{h} - \Delta_{\mu}\bigr)
\;+\;
\mathbf{W}_2\,\Delta_{\mu}.
\]

The second term is a constant vector, so we can shift it into the bias:
\[
\mathbf{b}_2' \;=\; \mathbf{b}_2 \;+\; \mathbf{W}_2\,\Delta_{\mu},
\]
which results in an equivalent output:
\begin{equation}
\label{eq:mean-shift}
\mathbf{y}
\;=\;
\mathbf{W}_2\,\bigl(\mathbf{h} - \Delta_{\mu}\bigr)
\;+\;
\mathbf{b}_2'.
\end{equation}

Because for each pruned index \(j\), we have replaced the activations $h_i$ by their means \(h_j = \mu_j\), it follows that \((\mathbf{h}-\Delta_{\mu})_j = 0\). Consequently, we can both drop the row \(j\) of \(\mathbf{W}_1\) as well as the column \(j\) of \(\mathbf{W}_2\) without affecting the computation, reducing the hidden dimension by \(|\mathcal{P}|\).

\paragraph{Why Use Trained Networks?}
Our approach uses the fact that \(\mathbf{W}_1, \mathbf{W}_2, \mathbf{b}_1, \mathbf{b}_2\) are constants in trained networks. This allows the mean to be easily shifted into the bias of the next layer, which in turn allows the low-variance neurons to be removed without requiring extensive additional training. Only minor fine-tuning is sufficient to adjust the remaining weights for best performance.

\section{Experiments}
\label{sec:experiments}

\begin{table*}[t]
	\centering
\begin{tabularx}{\textwidth}{
		p{2.1cm} 
		>{\centering\arraybackslash}X 
		>{\centering\arraybackslash}p{2.5cm} 
		>{\centering\arraybackslash}X 
		>{\centering\arraybackslash}p{2.5cm}
		>{\centering\arraybackslash}X 
		>{\centering\arraybackslash}p{2.5cm} 
		>{\centering\arraybackslash}p{2.5cm}
	}
	\toprule
	\textbf{Model} & \multicolumn{2}{c}{\textbf{MACs (G)}} & \multicolumn{2}{c}{\textbf{Parameters (M)}} & \multicolumn{3}{c}{\textbf{Top-1 Acc. (\%)}} \\ 
	\cmidrule(lr){2-3} \cmidrule(lr){4-5} \cmidrule(lr){6-8}
	(Pruning Rate) & \textbf{Full} & \textbf{VBP} & \textbf{Full} & \textbf{VBP} & \textbf{Full} & \textbf{Ret.} & \textbf{VBP} \\ 
	\midrule
	DeiT-T (45\%)		      	& 1.26 	& 0.94 (-25.16\%)	& 5.72 	& 4.12 (-27.97\%) & 72.02 	& 49.77 (69.13\%)        	& 70.08 (97.33\%)  \\
	DeiT-S (50\%)		      	& 4.61 	& 3.21 (-30.37\%)	& 22.05	& 14.96 (-32.15\%) & 79.70 	& 64.44 (80.85\%)        	& 78.62 (98.64\%)  \\
	DeiT-B (55\%)		      	& 17.58  		 	& 11.44 (-34.93\%)	& 86.57 	& 55.40 (-36.01\%) & 81.73 		& 57.58 (70.48\%)        	& 80.67 (98.74\%)  \\
	\midrule
	DeiT-B (20\%)		      	& 17.58  		 	& 15.35 (-12.68\%)	& 86.57 	& 75.24 (-13.09\%) & 81.73 		& 80.87 (98.98\%)        	& 81.76 (100.07\%)  \\
	\midrule
	Swin-T (45\%)		      	& 4.50 		& 3.23 (-28.22\%)		& 28.29 	& 21.31 (-24.67\%) & 80.91 	& 55.12 (68.13\%)       		& 79.41 (98.15\%)  \\  
	Swin-S (50\%)		      	& 8.76 	& 5.63 (-32.19\%)	& 49.61	& 35.02 (-29.41\%) & 83.04 	& 67.01 (80.70\%)       	& 81.86 (98.58\%)   \\  
	Swin-B (55\%)		      	& 15.46 	& 10.22 (-33.89\%)	& 87.77 	& 56.29 (-35.87\%) & 84.71 	& 62.61 (73.91\%)       	& 83.61 (98.70\%)   \\  
	\midrule
	Swin-B (20\%)		      	& 15.46 	& 13.58 (-12.16\%)	& 87.77 	& 76.42 (-12.93\%) & 84.71 	& 83.90 (99.04\%)       	& 84.67 (99.95\%)   \\  
	\bottomrule
\end{tabularx}

	\caption{Results comparing the \textbf{full} baseline model with our \textbf{VBP} model using four metrics: \textbf{MACs}: computational operations, measured in billions of operations; and \textbf{Parameters}: the total model size in millions of parameters; \textbf{Accuracy Retention (Ret.)}: retained accuracy after pruning, before fine-tuning; and \textbf{Final Accuracy}: accuracy after fine-tuning. Our method achieves competitive accuracy with significant reductions in MACs and parameters after fine-tuning, and even right after pruning, when reducing the pruning rate to 20\%.}
	\label{tab:main-results}
\end{table*}

In this section, we apply our \gls{vbp} method to various transformers and similar architectures, namely \gls{deit}, \gls{swin}, and \gls{convnext} in the Tiny, Small, and Base sizes, all trained on the \gls{imagenet}~\citep{ialshid} dataset. 

We evaluate these models twice. Once immediately after pruning (before fine-tuning) to measure how much accuracy is retained, and the second time after a brief fine-tuning to measure the final accuracy. Fine-tuning is performed over merely 10 epochs using knowledge-distillation from the unmodified base model. We use the AdamW optimizer with an initial learning rate of 1.5e-5, decayed by a cosine-annealing scheduler, and a batch size of 32. As regularization, we include a weight decay of 0.01. All experiments are performed on NVIDIA H200 Tensor Core GPUs.

\subsection{Main Results}

As is common with various pruning techniques, the pruning rate is adapted for the size of the models. Larger models typically exhibit higher redundancy for similar tasks, and therefore allow for more pruning while maintaining accuracy. We report our results for \gls{deit} and \gls{swin} in \cref{tab:main-results} and \cref{tab:runtime-results}. Note that the pruning rate is applied to the \glspl{mlp} only and does not indicate the total reduction in model size.
\begin{itemize}
	\item For the base-sized models, we prune all \glspl{mlp} globally using a pruning rate of $55\%$ keeping $99\%$ of the original performance after fine-tuning, reducing the total model size of \gls{deit}-Base, \gls{swin}-Base both by $36\%$. This corresponds to $35\%$, and $34\%$ fewer MACs respectively and provides a speed-up of $1.44\times$ and $1.30\times$ times.
	\item The small-sized models have notably fewer parameters than the base models and are thus pruned $50\%$ again reaching $99\%$ of the original performance. This leads to a DeiT-Small size reduction of $32\%$ and Swin-Small reduction of $29\%$ with $30\%$ and $32\%$ MACs savings respectively for a total speed-up factor of $1.34\times$ and $1.29\times$.
	\item Finally, for the \gls{deit}- and \gls{swin}-Tiny models, we apply a lower pruning rate of $45\%$, achieving $97\%$ and $98\%$ final performance with $28\%$ and $25\%$ reductions in model size. This translates to $25\%$ and $28\%$ fewer MACs, along with speed-ups of $1.17\times$ and $1.20\times$ respectively.
	\item For the Base-Models, we demonstrate that when globally pruning the \glspl{mlp} by $20\%$ the accuracy retention directly after the one-shot structured pruning reaches $99\%$ of the unpruned performance without any fine-tuning. This allows for an off-the-shelf $13\%$ size and MACs reduction and a speed-up of $1.11\times$ for \gls{deit}-Base and a similar reductions for \gls{swin}-Base
\end{itemize}
All further pruning rates ranging from $5\%$ to $50\%$ for all \gls{deit} Models can be found in \cref{sec:pruning-rates}.

\begin{table}[t] 
	\centering
	\begin{tabularx}{\columnwidth}{
			p{2.5cm}
			>{\centering\arraybackslash}X
			>{\centering\arraybackslash}X
			>{\centering\arraybackslash}X			
			>{\centering\arraybackslash}X
		}
		\toprule	
		\textbf{Model} & \textbf{Tiny}  				& \textbf{Small} 			& \textbf{Base} & \textbf{Base} \\
		(Pruning Rate) &  (45\%) &  (50\%) &  (55\%) &  (20\%) \\
		\midrule
		DeiT         				& 1.17         					& 1.34 						& 1.44 & 1.11 \\
		Swin         				& 1.20	          				& 1.27 						& 1.30 & 1.10\\
		\bottomrule
	\end{tabularx}
	\caption{Speed-ups of the pruned models relative to the baseline for different sizes and pruning rates.}
	
	\label{tab:runtime-results}
\end{table}

\subsection{Performance Comparison}

\begin{table}[t] 
	\centering
	\begin{tabularx}{\columnwidth}{
			p{2.5cm}
			>{\centering\arraybackslash}X 
			>{\centering\arraybackslash}X 
			>{\centering\arraybackslash}X 
			>{\centering\arraybackslash}X 
		}
		\toprule
		\textbf{Model} & \textbf{MACs} & \textbf{Param.} & \multicolumn{2}{c}{\textbf{Top-1 Acc. (\%)}} \\ 
		\cmidrule(lr){4-5}
		& \textbf{(G)} & \textbf{(M)} & \textbf{Ret.} & \textbf{Final} \\ 
		\midrule
		\rowcolor{gray!20} 
		DeiT-T         				& 1.26  					& 5.72 			& --         	& 72.02 \\
		Magnitude         			& 0.91   					& 3.94 			& 3.56       	& 68.49 \\
		SNIP	         			& 0.91   					& 3.94 			& 24.89      	& 69.10 \\
		VBP (ours)    				& 0.91   		 			& 3.94 			& \textbf{39.58}  & \textbf{70.61} \\
		\midrule
		\rowcolor{gray!20} 
		DeiT-S         				& 4.61  					& 22.05 		& --         	& 79.70 \\
		Magnitude         			& 3.21   					& 14.96 		& 4.05       	& 76.55 \\
		SNIP	         			& 3.21   					& 14.96 		& 52.39      	& 77.27 \\
		VBP (ours)    				& 3.21  				  	& 14.96 		& \textbf{64.44}  & \textbf{78.62} \\
		\midrule
		\rowcolor{gray!20} 
		DeiT-B         				& 17.58  					& 86.57 		& --         	& 81.73 \\
		Magnitude         			& 12.0   					& 58.24 		& 0.37       	& 78.88 \\
		SNIP	         			& 12.0   					& 58.24 		& 53.24      	& 80.40 \\
		VBP (ours)    				& 12.0	    				& 58.24 		& \textbf{66.40}  & \textbf{80.99} \\
		\bottomrule
	\end{tabularx}

	\caption{Comparison of different one-shot pruning scores in structured pruning to our \textbf{VBP} approach at fixed 50\% pruning rates. For constant \textbf{MACs} and \textbf{parameters}, our approach retains more accuracy (\textbf{Ret.}) and reaches a higher \textbf{final} accuracy compared to other pruning approaches.}
	
	\label{tab:comparison-results}
\end{table}

\begin{table*}[t] 
	\centering
\begin{tabularx}{\textwidth}{
		p{2.5cm}
		>{\centering\arraybackslash}X 
		>{\centering\arraybackslash}X 
		>{\centering\arraybackslash}X 
		>{\centering\arraybackslash}X 
		>{\centering\arraybackslash}X 
	}
	\toprule
	\textbf{Model} & \textbf{MACs (G)} & \textbf{Param. (M)} & \textbf{Speed-Up} & \multicolumn{2}{c}{\textbf{Top-1 Acc. (\%)}} \\ 
	\cmidrule(lr){5-6}
	& & & & \textbf{Ret.} & \textbf{Final} \\ 
	\midrule
	\rowcolor{gray!20} 
	DeiT-B         				& 17.58  					& 86.57 	& --  & --         	& 81.73 \\
	ToMe-24         			& 6.02  					& 86.57 	& 2.15  & 35.85       	& 75.74 \\
	ToMe-14 \& VBP	         	& 6.08   					& \textbf{58.24} 	& 2.05  & 60.53      	& \textbf{80.09} \\			
	\midrule
	ToMe-20	         			& 7.14			 		  	& 86.57 	& 1.84  & 66.31      	& 78.66 \\				
	ToMe-12 \& VBP	         	& 6.90   					& \textbf{58.24} 	& 1.83  & 62.09	    & \textbf{80.29} \\
	\midrule
	ToMe-18         			& 7.88   					& 86.57 	& 1.75  & 73.79	    & 79.97 \\
	ToMe-10 \& VBP	         	& 7.73   					& \textbf{58.24} 	& 1.70  & 63.72	    & \textbf{80.45} \\				
	\bottomrule
\end{tabularx}

\caption{Comparison of ToMe for different token reduction rates of 24, 20, and 18, with the hybrid combination of ToMe using token reduction rates of 14, 12, and 10, and \textbf{VBP} using 50\% pruning rates on top. Our hybrid method achieves significantly higher \textbf{final} accuracies throughout while also providing reductions in \textbf{parameters}, for similar \textbf{MACs} and \textbf{speed-ups}. The hybrid model thereby reaches speed-ups of up to 2.05$\times$ while maintaining competitive performance.}
\label{tab:performance}
\end{table*}

\begin{table}[t] 
	\centering
\begin{tabularx}{\columnwidth}{
		p{2.5cm}
		>{\centering\arraybackslash}X 
		>{\centering\arraybackslash}X 
		>{\centering\arraybackslash}X 
		>{\centering\arraybackslash}X 
	}
	\toprule
	\textbf{Model} & \textbf{Epochs} & \textbf{Param.} & \multicolumn{2}{c}{\textbf{Top-1 Acc. (\%)}} \\ 
	\cmidrule(lr){4-5}
	& & (M) & \textbf{Ret.} & \textbf{Final} \\ 
	\midrule
	NViT (50 Ep.)         			& 50          				& 56.37		 			& 81.13 						& 82.18   						 \\
	\midrule
	NViT (1 Ep.)       			& 1         				& 56.85 	 			& 69.10 						& 81.92  						 \\
	VBP (ours)	       				& 1							& 56.18 				& \textbf{72.37}          		& \textbf{82.32} 				 \\
	\bottomrule
\end{tabularx}
\caption{Comparison with NViT using two pruning durations of 50 epochs (as originally proposed) and 1 epoch (similar to our VBP approach), both followed by 10 epochs of fine-tuning. Our method achieves higher accuracy immediately after pruning, before fine-tuning (\textbf{Ret.}) compared to NViT with the same pruning duration, and reaches a higher \textbf{final} accuracy than the 50-epoch NViT model after fine-tuning for 10 epochs.}
\label{tab:NViT}
\end{table}

To facilitate direct comparisons of all models with different sizes across all experiments, we fix the pruning rate at $50\%$ for all further experiments. 

We demonstrate the effectiveness of our selection criterion, by benchmarking against two other common and well-established one-shot-pruning baselines, namely Magnitude and SNIP, adapted for structured pruning. 
For a fixed pruning rate of $50\%$ using all selection criteria, this produces identical parameter counts of $3.94$, $14.96$, and $58.24$ million parameters, as well as fixed $0.91$, $3.21$, and $12.00$ GMACs for the three model sizes. Our results indicate that \gls{vbp} provides higher accuracy retention compared to the baselines, improving \gls{deit}-Base by $13.16\%$, \gls{deit}-Small by $12.05\%$, and \gls{deit}-Tiny by $14.69\%$ percentage points compared to SNIP. This difference in accuracy retention yields a final fine-tuned performance that is $0.59\%$ to $1.51\%$ percentage points higher after fine-tuning.

\subsection{\gls{sota} Pruning Methods}

To further validate our approach, we compare against the \gls{sota} structured transformer pruning method NViT (CVPR'23). Specifically, we apply our method to the NViT Base model (DeiT-Base distilled from RegNetY-160) and run NViT with a latency target of $70\%$, pruning  once for 50 epochs (as per the published results) and once for one epoch (similar to our setup), with both scenarios followed by ten epochs of fine-tuning using the NViT settings. In a similar pruning setting, VBP achieves a $0.4\%$ percentage points higher final accuracy, for similar number of parameters.

Finally, we demonstrate the orthogonality of VBP to token-pruning methods by applying our method on top of the \gls{sota} token-merging approach, \gls{tome} (ICLR'23). We observe that the accuracy retention drops relative to the baseline remains consistent for both the \gls{vbp}-pruned base \gls{deit} model in \cref{tab:main-results} and the \gls{vbp}- \& \gls{tome}-pruned model. This relative consistency also persists throughout fine-tuning, indicating that our method operates orthogonally to \gls{tome} in terms of performance.

Regarding the MAC savings it is important to note that \gls{tome} progressively reduces the throughput of tokens processed across the layers. Later layers therefore process less tokens, making pruning less effective. The relative MAC savings from our hybrid method therefore decrease as the network depth increases. Nonetheless, \gls{vbp} additionally reduces the total parameter count, which cannot be achieved by simply increasing the \gls{tome} token reduction rate.  

In fact, increasing the reduction rate to get an approximate $2\times$ speed-up yields similar MAC savings but reduces both the accuracy retention as well as the final accuracy significantly. Instead applying  \gls{vbp} on top of \gls{tome} with a smaller reduction rate, allows more than $2\times$ speed-ups while still maintaining $98\%$ of the original accuracy with additional parameter savings.

\section{Analysis}
\label{sec:ablation}

\begin{table}[t] 
	\centering
	\begin{tabularx}{\columnwidth}{
			>{\centering\arraybackslash}X
			>{\centering\arraybackslash}X
			>{\centering\arraybackslash}X
			>{\centering\arraybackslash}X
		}
		\toprule
		\textbf{Variance-} & \textbf{Mean-} & \multicolumn{2}{c}{\textbf{Top-1 Acc. (\%)}} \\ 
		\cmidrule(lr){3-4} 
		\textbf{Based} & \textbf{Shift} & \textbf{Ret.} & \textbf{Final}  \\ 
		\midrule
		\xmark & \cmark & 55.19 & 80.23 \\ 
		\cmark & \xmark & 26.04 & 80.62 \\ 
		\cmark & \cmark & \textbf{66.40} & \textbf{80.99} \\ 
		\bottomrule
	\end{tabularx}
	\caption{Ablation study results showing the impact of Variance-Based Pruning, and Mean-Shift Compensation on the retained accuracy after pruning (\textbf{Ret.}), as well as the \textbf{final}: accuracy after fine-tuning. Both steps together yield the highest accuracy of 80.99\%.}
	\label{tab:ablation-results}
\end{table}

\begin{figure}[t] 
	\centering
	\includegraphics[width=\columnwidth]{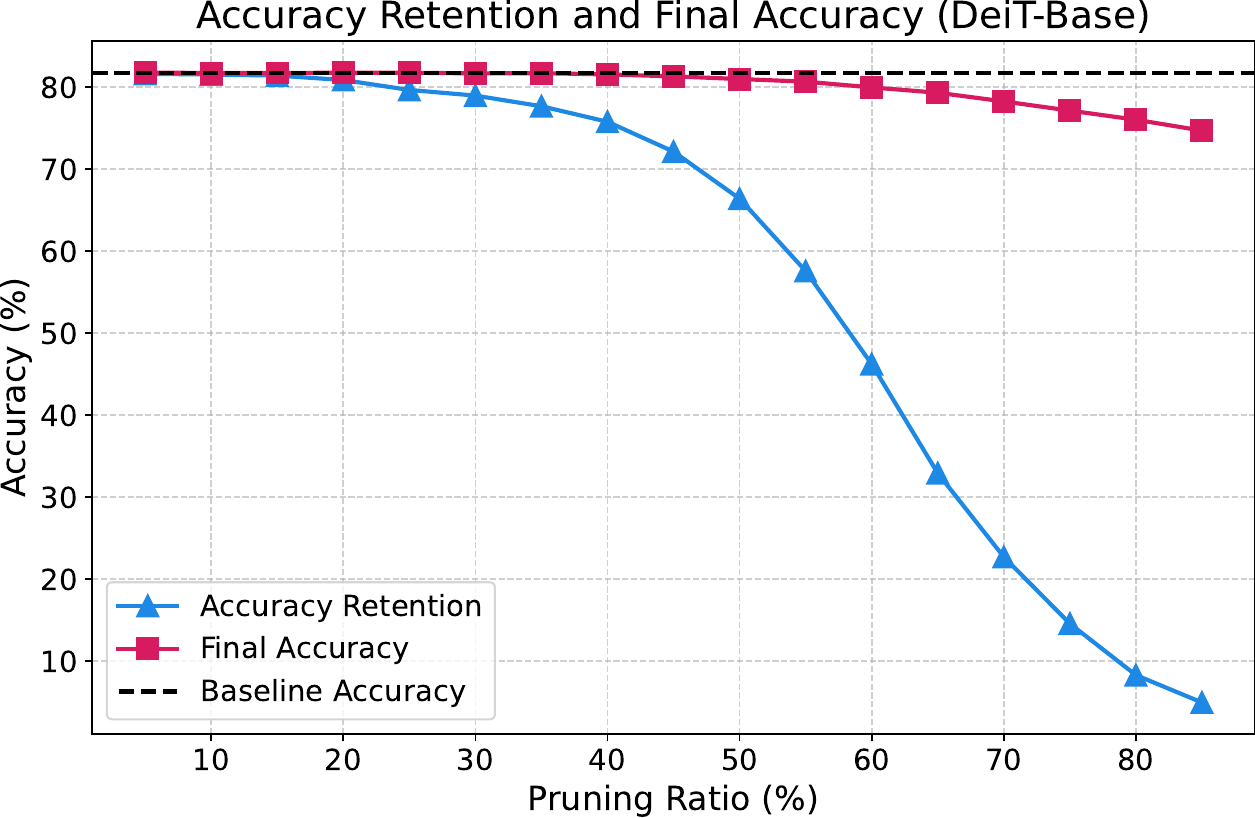}
	\caption{Accuracy retention before fine-tuning and final accuracy after 10 epochs of fine-tuning for different pruning rates applied to \gls{deit}-Base.}
	\label{fig:pruning-rates}
\end{figure}

\begin{figure}[t] 
	\centering
	\includegraphics[width=\columnwidth]{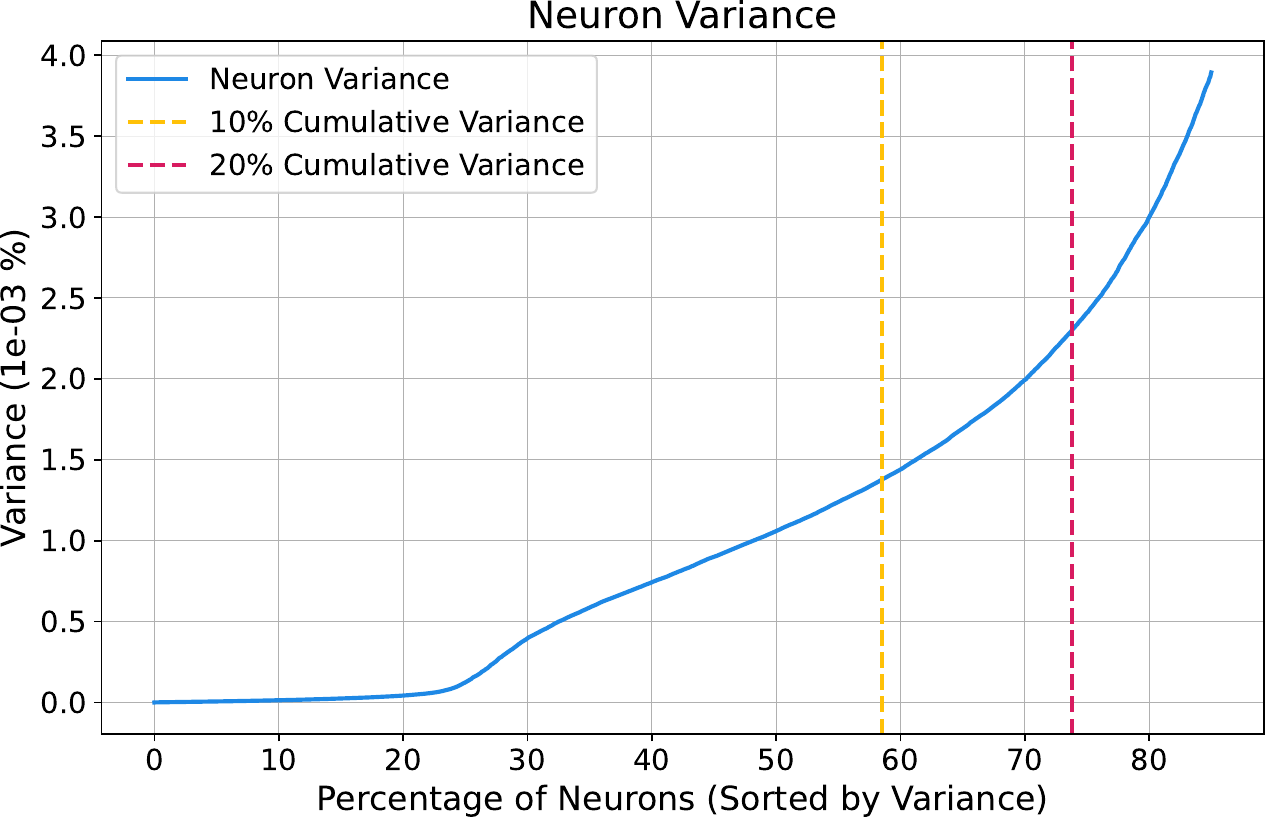}
	\caption{Activation variance for all neurons in the hidden \gls{mlp} layers throughout the network and marked cumulative variances.}
	\label{fig:variance-distribution}
\end{figure}

\begin{table}[t] 
	\centering
\begin{tabularx}{\columnwidth}{
		p{2.7cm}
		>{\centering\arraybackslash}X 
		>{\centering\arraybackslash}X 
		>{\centering\arraybackslash}X 
		>{\centering\arraybackslash}X 
	}
	\toprule
	\textbf{Model} & \textbf{MACs} & \textbf{Param.} & \multicolumn{2}{c}{\textbf{Top-1 Acc. (\%)}} \\ 
	\cmidrule(lr){4-5}
	& \textbf{(G)} & \textbf{(M)} & \textbf{Ret.} & \textbf{Final} \\ 
	\midrule
	Pre-Act. (eq. \citep{vpplmvtnv})  & 12.0   & 58.24  & 0.43  & 77.92 \\
	Post-Act. (ours)          & 12.0   & 58.24  & \textbf{66.40}  & \textbf{80.99} \\
	\bottomrule
\end{tabularx}

\caption{Comparison of different locations for the activation statistics computation in the network: before (\textbf{Pre-Act.}) as well as after the activation function (\textbf{Post-Act.}). For the same \textbf{MACs} and \textbf{parameters}, using pruning rates of 50\%, gathering statistics post-activation significantly outperforms pre-activation in both accuracy retention (\textbf{Ret.}) as well as the \textbf{final} accuracy.}
	
	\label{tab:pre-post-activation}
\end{table}

To assess the contribution of each individual component of our approach, we perform an ablation study in which we systematically modify the pruning criterion and the Mean-Shift Compensation. We report our results in \cref{tab:ablation-results}. 
\\
We first remove the variance-based pruning selection while still shifting the mean activation to the next bias. Removing the Variance-Based Pruning alone reduces the final accuracy by $0.76\%$ percentage points.

We also remove the Mean-Shift Compensation while pruning using the variance-based criterion. This improves the final accuracy by $0.39\%$ compared to using only the Mean-Shift Compensation. 
 
However, the best results are achieved when both Variance-Based Pruning and Mean-Shift Compensation are used together. As discussed in \cref{sec:optimality}, replacing the removed activations with their means is optimal when pruning based on variance. When both components are combined, the immediate accuracy retention increases by $11.21\%$ percentage points, ultimately improving the fine-tuned accuracy by an additional $0.37\%$.
This highlights the significance of both components.

\subsection{Sensitivity of Variance Thresholding}

We further examine the stability of the activation variance criterion by applying different pruning rates to the \gls{deit}-Base model, and plotting the accuracy retention and final accuracy in \cref{fig:pruning-rates}. We observe that for smaller pruning rates (up to around 25\%), the accuracy retention is high enough to allow for off-the-shelf deployment without any fine-tuning. 

Beyond these rates, the accuracy begins to degrade more noticeably, reflecting the non-uniform distribution of neuron variances, as seen in \cref{fig:variance-distribution}. 
Notably, to account for $10\%$ of the cumulative variance across all neuons, nearly $60\%$ of the lowest-variance neurons are needed. This is a disproportionately large share compared to the next $15\%$ of neurons that also account for $10\%$ of the total variance in a layer. Consequently, more aggressive pruning leads to an increasingly faster reduction in the total variance and thus expressiveness of the network. Nevertheless, significant performance issues arise only at very high pruning levels, validating the feasibility of our method over a broad range of pruning rates (see \cref{tab:pruning-rates}).

\begin{table*}[t]
	\centering
\begin{tabularx}{\textwidth}{
		p{2cm} 
		>{\centering\arraybackslash}X 
		>{\centering\arraybackslash}p{2.5cm} 
		>{\centering\arraybackslash}X 
		>{\centering\arraybackslash}p{2.5cm} 
		>{\centering\arraybackslash}X 
		>{\centering\arraybackslash}p{2.5cm} 
		>{\centering\arraybackslash}p{2.5cm}
	}
	\toprule
	\textbf{Model} & \multicolumn{2}{c}{\textbf{MACs (G)}} & \multicolumn{2}{c}{\textbf{Parameters (M)}} & \multicolumn{3}{c}{\textbf{Top-1 Acc. (\%)}} \\ 
	\cmidrule(lr){2-3} \cmidrule(lr){4-5} \cmidrule(lr){6-8}
	& \textbf{Full} & \textbf{VBP} & \textbf{Full} & \textbf{VBP} & \textbf{Full} & \textbf{Retention} & \textbf{VBP} \\ 
	\midrule
	ConvNeXt-T		      	& 4.47 	& 2.96 (-33.8\%)	& 28.59 	& 12.61 (-55.9\%) & 82.90 	& 16.80 (20.3\%)       	& 81.30 (98.1\%)   \\  
	ConvNeXt-S		      	& 8.71 	& 5.11 (-41.3\%)	& 50.22	& 23.50 (-53.2\%) & 84.57 	& 30.92 (36.6\%)       	& 82.82 (97.9\%)   \\  
	ConvNeXt-B		      	& 15.38 	& 8.91 (-42.1\%)	& 88.59 	& 41.32 (-53.4\%) & 85.51 	& 57.10 (66.8\%)       	& 83.40 (97.6\%)   \\  
	\bottomrule
\end{tabularx}

\caption{Results comparing the \textbf{full} ConvNeXt baseline models with the \textbf{VBP} models using a pruning rate of 50\%. The \textbf{MACs} are reduced by up to 42\% and \textbf{parameters} by over 50\% while reaching competitive \textbf{final} accuracies of 98\% original performance.}

	\label{tab:convnext-main-results}
\end{table*}

\begin{table}[t] 
	\centering
	\begin{tabularx}{\columnwidth}{
			p{2.5cm}
			>{\centering\arraybackslash}X
			>{\centering\arraybackslash}X
			>{\centering\arraybackslash}X
		}
		\toprule	
		\textbf{Model}      		& \textbf{Tiny} 				& \textbf{Small} 			& \textbf{Base} \\
		\midrule
		ConvNeXt	         		& 1.28          				& 1.42 						& 1.49  \\
		\bottomrule
	\end{tabularx}
\caption{Speed-ups of the pruned models relative to the baseline for different-sized ConvNeXt models.}

	\label{tab:convnext-runtime-results}
\end{table}

\subsection{Application Pre- vs. Post-Activation}

While \cref{sec:optimality} shows that our mean-replacement strategy can be grounded in theory for any distribution, the question remains whether the variance should be measured before or after the activation function. The Central Limit Theorem confirms that pre-activation sums (composed of independent weight contributions) tend to approximate a normal distribution. However, once the nonlinearity \(\sigma\) is applied, this distribution changes. Consequently, the pre-activation variance does not necessarily correlate with the neuron importance for retaining accuracy in an already trained model.

For instance, highly varying negative activations are compressed into range of about $(-0.2, 0)$ by GeLU, reducing their variance.  As a result, such a neuron would be less likely to be pruned in a pre-activation setting and more likely to be pruned post-activation (see \cref{sec:pre-post-act}).

To validate this experimentally, we tested an alternative design in which pruning decisions are made based on the pre-activation variance in \cref{tab:pre-post-activation}. While this approach has been successfully applied when pruning during training \citep{vpplmvtnv}, we find that in a trained network setting, it significantly degrades performance. Therefore, measuring variance \emph{post}-activation better captures neuron importance in already trained models, and helps maintain model performance despite structural modifications.

\subsection{Other Transformer-Like Architectures}

While our method is broadly applicable to any model containing \gls{mlp} layers, network architectures that have more and larger \glspl{mlp} benefit more from \gls{vbp}. We therefore apply our pruning on \gls{convnext} as a transformer-like architecture, that relies heavily on \gls{mlp} layers, to further evaluate the generilizability of our method.

We summarize our performance comparisons for \gls{convnext} in \cref{tab:convnext-main-results}. We achieve significant parameter reductions of over $50\%$ throughout all three sizes and reduced MACs by $34\%$ for \gls{convnext}-Tiny and over $40\%$ for the larger sizes, while maintaining $98\%$ of the original accuracy after 10 epochs of fine-tuning across the board. The corresponding speed-ups are listed in \cref{tab:convnext-runtime-results}, and reach up to $1.49\times$ for \gls{convnext}-Base.

Noteably, we observe significantly less accuracy retention throughout the smaller sized models compared to the true transformers. We attribute this to the fact that the \gls{convnext} architecture is built upon convolutions, which have significantly higher costs compared to their parameters. Consequently, since the \glspl{mlp} take over a higher overall percentage of the total capacity, removing a similar fraction of neurons causes a relatively higher drop in accuracy retention. By contrast, transformers typically have less parameters in their MLP relatively speaking, therby allowing for higher pruning rates while maintaining capacity.

\section{Conclusion}
\label{sec:conclusion}

We introduce \gls{vbp}, a one-shot pruning method, designed to remove neurons based on their statistical importance in a single pruning operation while minimizing accuracy loss by integrating their mean activations in a Mean-Shift Compensation step.

Our experiments demonstrate that \gls{vbp} retains model performance significantly better than existing structured pruning methods. By taking advantage of trained networks, our approach provides a favourable starting point for the subsequent finetuning, eliminating the need for extensive retraining and thereby making it practical for low-cost deployment. 

Furthermore, we show that \gls{vbp} operates orthogonally to \gls{sota} token pruning methods, allowing it to be seamlessly combined with approaches like \gls{tome} to achieve even greater computational savings.

We believe that the simplicity and efficiency of our approach contributes to the democratization of deep learning by enabling a wider reuse of trained networks and hope it inspires further research in the \gls{cv} community toward efficient model compression techniques.

{
    \small
    \bibliographystyle{ieeenat_fullname}
    \bibliography{main}
}
\appendix
\maketitlesupplementary
\setcounter{page}{1}

\begin{figure}[h] 
	\centering
	\includegraphics[width=\columnwidth]{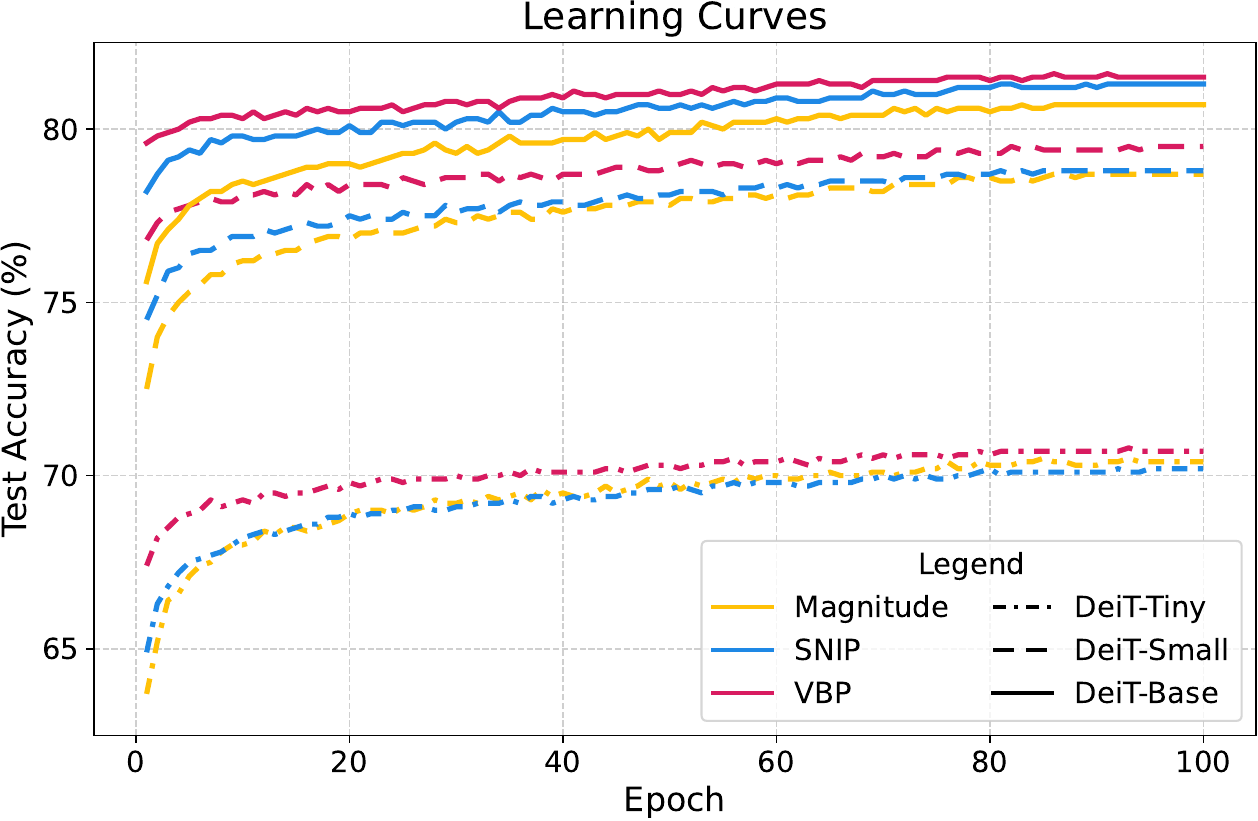}
	\caption{Learning curves over 100 epochs of fine-tuning after different structured pruning methods. VBP retains a performance lead over other methods throughout the entire training period.}
	\label{fig:learning-curves-all-methods}
\end{figure}

\begin{figure}[h] 
	\centering
	\includegraphics[width=\columnwidth]{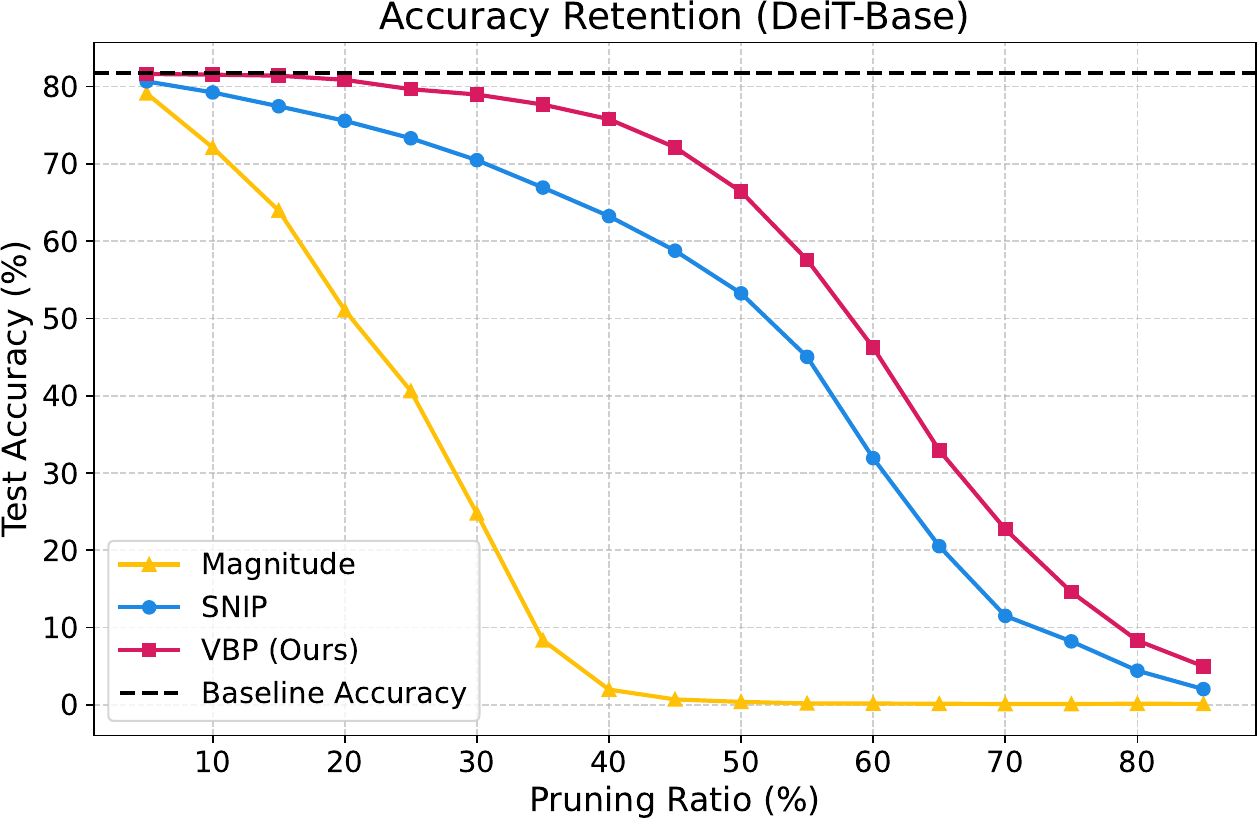}
	\caption{Accuracy retention across varying pruning ratios for different structured pruning methods applied to \gls{deit}-Base. VBP consistently retains more accuracy across all pruning levels.}
	\label{fig:accuracies-per-pruning-rate-all-methods}
\end{figure}

\begin{figure}[h] 
	\centering
	\includegraphics[width=\columnwidth]{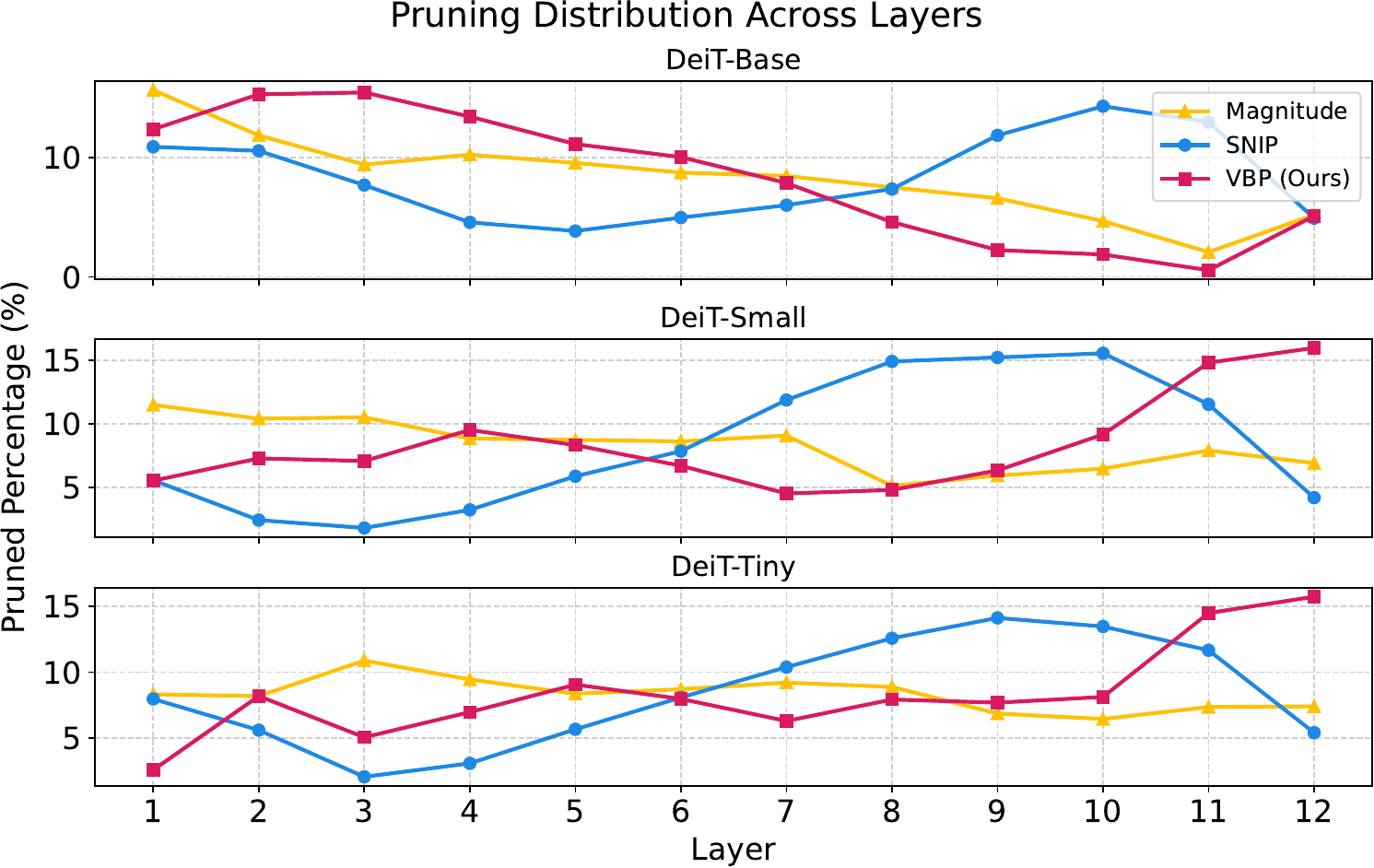}
	\caption{Layer-wise pruning distributions  for different structured pruning methods. VBP prunes more in early layers, contrasting with SNIP which increases pruning in deeper layers.}
	\label{fig:deit-layerwise-pruning-all-methods}
\end{figure}

\section{Additional Analysis}
\label{sec:additional-analysis}

\subsection{Longer Training and Learning Curves}

To validate the sustained performance of VBP throughout training, we extend training to 100 epochs and compare learning curves against other pruning methods adapted for structured pruning. While the performance gap narrows over time, VBP consistently outperforms alternatives throughout all 100 epochs, as shown in \cref{fig:learning-curves-all-methods}.

\subsection{Comparison at Different Pruning Ratios}

We further examine the stability of VBP compared to other pruning methods adapted for structured pruning. We apply varying pruning rates to the \gls{deit}-Base model and plot the accuracy retention. As shown in \cref{fig:accuracies-per-pruning-rate-all-methods}, VBP maintains a consistent advantage across all pruning levels, indicating its relative robustness to aggressive pruning.

\subsection{Pruning Distribution Across Layers}

We analyze how pruning decisions are distributed across layers. Interestingly, gradient-based methods such as SNIP tend to prune neurons in a pattern opposite to that of VBP, which focuses more on early layers. This difference is visualized in \cref{fig:deit-layerwise-pruning-all-methods}, suggesting that VBP’s strategy may lead to better feature preservation and downstream performance.

\onecolumn

\section{Performance on \gls{imagenet} for Different Pruning Rates}
\label{sec:pruning-rates}

\begin{table}[H]
	\label{tab:pruning-rates}	
	\centering
	\begin{tabularx}{\textwidth}{
			p{3cm} 
			>{\centering\arraybackslash}X 
			>{\centering\arraybackslash}X 
			>{\centering\arraybackslash}X 
			>{\centering\arraybackslash}X 
		}
		\toprule
		\textbf{Model} & \textbf{MACs (G)} & \textbf{Parameters (M)} & \multicolumn{2}{c}{\textbf{Top-1 Acc. (\%)}} \\ 
		\cmidrule(lr){4-5}
		(Pruning Rate) & & & \textbf{Retention} & \textbf{Final} \\ 
		\midrule
		\rowcolor{gray!20} 
		\textbf{DeiT-Base}   & 17.58 		   & 86.57  & -     			   & 81.73  			  \\ 
		5\%   & 17.02 (-3.19\%) & 83.73 (-3.28\%) & 81.63 (99.91\%)   & 81.79 (100.11\%) \\  
		10\%  & 16.46 (-6.37\%) & 80.90 (-6.55\%) & 81.54 (99.80\%)   & 81.72 (100.02\%) \\  
		15\%  & 15.91 (-9.50\%) & 78.07 (-9.82\%) & 81.41 (99.65\%)   & 81.73 (100.04\%) \\  
		20\%  & 15.35 (-12.68\%) & 75.24 (-13.09\%) & 80.87 (98.98\%)   & 81.76 (100.07\%) \\  
		25\%  & 14.79 (-15.87\%) & 72.40 (-16.37\%) & 79.66 (97.50\%)   & 81.76 (100.07\%) \\  
		30\%  & 14.23 (-19.06\%) & 69.57 (-19.64\%) & 78.97 (96.66\%)   & 81.68 (99.98\%) \\  
		35\%  & 13.67 (-22.24\%) & 66.74 (-22.91\%) & 77.67 (95.07\%)   & 81.71 (100.01\%) \\  
		40\%  & 13.12 (-25.37\%) & 63.90 (-26.19\%) & 75.78 (92.75\%)   & 81.55 (99.82\%) \\  
		45\%  & 12.56 (-28.56\%) & 61.07 (-29.46\%) & 72.14 (88.30\%)   & 81.32 (99.53\%) \\  
		50\%  & 12.00 (-31.74\%) & 58.24 (-32.72\%) & 66.40 (81.27\%)   & 80.99 (99.13\%) \\  
		\midrule
		\rowcolor{gray!20} 
		\textbf{DeiT-Small}   & 4.61 		   & 22.05  & -     			   & 79.70  			  \\ 
		5\%  & 4.47 (-3.04\%) & 21.34 (-3.22\%) & 78.63 (98.66\%)   & 79.67 (99.96\%) \\  
		10\%  & 4.33 (-6.07\%) & 20.63 (-6.44\%) & 78.32 (98.27\%)   & 79.63 (99.91\%) \\  
		15\%  & 4.19 (-9.11\%) & 19.92 (-9.66\%) & 77.65 (97.43\%)   & 79.65 (99.94\%) \\  
		20\%  & 4.05 (-12.15\%) & 19.22 (-12.83\%) & 76.90 (96.49\%)   & 79.65 (99.94\%) \\  
		25\%  & 3.91 (-15.18\%) & 18.51 (-16.05\%) & 76.13 (95.52\%)   & 79.63 (99.91\%) \\  
		30\%  & 3.77 (-18.22\%) & 17.80 (-19.27\%) & 74.98 (94.08\%)   & 79.52 (99.77\%) \\  
		35\%  & 3.63 (-21.26\%) & 17.09 (-22.49\%) & 73.49 (92.21\%)   & 79.30 (99.50\%) \\  
		40\%  & 3.49 (-24.30\%) & 16.38 (-25.71\%) & 71.44 (89.64\%)   & 79.18 (99.35\%) \\  
		45\%  & 3.35 (-27.33\%) & 15.67 (-28.93\%) & 68.42 (85.85\%)   & 78.90 (99.00\%) \\  
		50\%  & 3.21 (-30.37\%) & 14.96 (-32.15\%) & 64.44 (80.85\%)   & 78.62 (98.64\%) \\  
		\midrule
		\rowcolor{gray!20} 
		\textbf{DeiT-Tiny} & 1.26 		   & 5.72  & -     			   & 72.02  			  \\
		5\%  & 1.22 (-3.17\%) & 5.54 (-3.15\%) & 71.67 (99.54\%)   & 72.05 (100.07\%) \\  
		10\%  & 1.19 (-5.56\%) & 5.36 (-6.29\%) & 70.95 (98.54\%)   & 71.92 (99.89\%) \\  
		15\%  & 1.15 (-8.73\%) & 5.18 (-9.44\%) & 70.05 (97.29\%)   & 71.76 (99.67\%) \\  
		20\%  & 1.12 (-11.11\%) & 5.01 (-12.41\%) & 68.87 (95.65\%)   & 71.60 (99.44\%) \\  
		25\%  & 1.08 (-14.29\%) & 4.83 (-15.56\%) & 67.37 (93.57\%)   & 71.44 (99.22\%) \\  
		30\%  & 1.05 (-16.67\%) & 4.65 (-18.71\%) & 64.76 (89.94\%)   & 71.20 (98.89\%) \\  
		35\%  & 1.01 (-19.84\%) & 4.48 (-21.68\%) & 61.12 (84.89\%)   & 70.86 (98.42\%) \\  
		40\%  & 0.98 (-22.38\%) & 4.30 (-24.83\%) & 55.64 (77.28\%)   & 70.55 (97.99\%) \\  
		45\%  & 0.94 (-25.16\%) & 4.12 (-27.97\%) & 49.77 (69.13\%)   & 70.08 (97.33\%) \\  
		50\%  & 0.91 (-27.94\%) & 3.94 (-31.12\%) & 39.58 (54.97\%)   & 69.70 (96.81\%) \\  
		
		\bottomrule
	\end{tabularx}
	\caption{Comparison evaluating four metrics on \gls{imagenet}~\citep{ialshid}: \textbf{MACs}: computational operations, measured in billions of operations; and \textbf{Parameters}: the total model size in millions of parameters; \textbf{Accuracy Retention (Ret.)}: retained accuracy after pruning, before fine-tuning; and \textbf{Final Accuracy}: accuracy after fine-tuning. Our method achieves competitive accuracy with significant reductions in MACs and parameters and allows off-the-shelf deployment for pruning rates up to 20\%.}
\end{table}

\clearpage

\section{Performance on \gls{cifar100} for Different Pruning Rates}
\label{sec:cifar100}

\begin{table}[H]
	\label{tab:cifar100}	
	\centering
\begin{tabularx}{\textwidth}{
		p{2.5cm} 
		>{\centering\arraybackslash}X 
		>{\centering\arraybackslash}X 
		>{\centering\arraybackslash}X 
		>{\centering\arraybackslash}X 
	}
	\toprule
	\textbf{Model} & \textbf{MACs (G)} & \textbf{Params (M)} & \multicolumn{2}{c}{\textbf{Top-1 Acc. (\%)}} \\ 
	\cmidrule(lr){4-5}
	(Pruning Rate) & & & \textbf{Retention} & \textbf{Final} \\ 
	\midrule
	\rowcolor{gray!20} 
	\textbf{DeiT-Base}   & 17.58 		   & 86.57  & -     			   & 88.23  			  \\ 
	30\%   & 14.23 (-19.06\%) & 69.57 (-19.64\%) & 84.83 (96.15\%)   & 88.07 (99.82\%) \\  
	40\%   & 13.12 (-25.37\%) & 63.90 (-26.19\%) & 82.21 (93.18\%)   & 87.53 (99.21\%) \\  
	50\%   & 12.00 (-31.74\%) & 58.24 (-32.72\%) & 77.45 (87.78\%)   & 87.00 (98.61\%) \\  
	60\%   & 10.88 (-38.11\%) & 52.57 (-39.27\%) & 54.60 (61.88\%)   & 85.98 (97.45\%) \\  
	70\%   & 10.33 (-41.24\%) & 49.74 (-42.54\%) & 32.58 (36.93\%)   & 84.84 (96.16\%) \\  
	\midrule
	\rowcolor{gray!20} 
	\textbf{DeiT-Small}   & 4.61 		   & 22.05  & -     			   & 85.43  			  \\ 
	30\%   & 3.77 (-18.22\%) & 17.80 (-19.27\%) & 71.88 (84.14\%)   & 85.28 (99.82\%) \\  
	40\%   & 3.49 (-24.30\%) & 16.38 (-25.71\%) & 68.41 (80.08\%)   & 85.48 (100.06\%) \\  
	50\%   & 3.21 (-30.37\%) & 14.96 (-32.15\%) & 60.99 (71.39\%)   & 84.94 (99.43\%) \\  
	60\%   & 2.93 (-36.44\%) & 13.55 (-38.55\%) & 51.17 (59.90\%)   & 83.65 (97.92\%) \\  
	70\%   & 2.65 (-42.52\%) & 12.13 (-44.99\%) & 33.19 (38.85\%)   & 82.08 (96.08\%) \\  
	\midrule
	\rowcolor{gray!20} 
	\textbf{DeiT-Tiny} & 1.26 		   & 5.72  & -     			   & 80.50  			  \\
	30\%   & 1.05 (-16.67\%) & 4.65 (-18.71\%) & 70.94 (88.12\%)   & 80.15 (99.57\%) \\  
	40\%   & 0.98 (-22.22\%) & 4.30 (-24.83\%) & 62.10 (77.14\%)   & 79.97 (99.34\%) \\  
	50\%   & 0.91 (-27.78\%) & 3.94 (-31.12\%) & 46.01 (57.16\%)   & 78.52 (97.54\%) \\  
	60\%   & 0.84 (-33.33\%) & 3.59 (-37.24\%) & 24.68 (30.66\%)   & 77.20 (95.90\%) \\  
	70\%   & 0.77 (-38.89\%) & 3.23 (-43.53\%) & 9.33 (11.59\%)   & 74.11 (92.06\%) \\  
	\bottomrule
\end{tabularx}
\caption{Comparison evaluating four metrics on \gls{cifar100}~\citep{lmloffti}: \textbf{MACs}: computational operations, measured in billions of operations; \textbf{Parameters}: the total model size in millions of parameters; \textbf{Accuracy Retention (Ret.)}: retained accuracy after pruning, before fine-tuning; and \textbf{Final Accuracy}: accuracy after fine-tuning.}
\end{table}

\clearpage

\section{Runtime Performance for Reported Models}
\begin{table}[h]
	\centering
	\label{tab:runtime-hardware-comparison}
	\begin{tabularx}{\textwidth}{
		p{3cm} 
		>{\centering\arraybackslash}p{1.3cm}
		>{\centering\arraybackslash}p{1.3cm}
		>{\centering\arraybackslash}X 
		>{\centering\arraybackslash}p{1.3cm} 
		>{\centering\arraybackslash}p{1.3cm} 
		>{\centering\arraybackslash}X 
		>{\centering\arraybackslash}p{1cm} 
		>{\centering\arraybackslash}p{1cm} 
		>{\centering\arraybackslash}X 
		}
    \toprule
	\textbf{Model} &
	\multicolumn{3}{c}{\textbf{H200 (GPU)}} &
	\multicolumn{3}{c}{\textbf{T4 (GPU)}} &
	\multicolumn{3}{c}{\textbf{E5 (CPU)}} \\
	\cmidrule(lr){2-4}
	\cmidrule(lr){5-7}
	\cmidrule(lr){8-10}
	(Pruning Rate) & \multicolumn{2}{c}{\textbf{Time (ms)}} & \textbf{Speed}& 
	\multicolumn{2}{c}{\textbf{Time (ms)}} & \textbf{Speed}& 
	\multicolumn{2}{c}{\textbf{Time (ms)}} & \textbf{Speed}\\
	\cmidrule(lr){2-3}
	\cmidrule(lr){5-6}
	\cmidrule(lr){8-9}
	& \textbf{Full} & \textbf{VBP} & \textbf{Up}
	& \textbf{Full} & \textbf{VBP} & \textbf{Up}
	& \textbf{Full} & \textbf{VBP} & \textbf{Up} \\
	\midrule
		DeiT-T (45\%) &
		3.64ms & 3.10ms & 1.17$\times$ &
		33.17ms & 28.36ms & 1.17$\times$ &
		0.53s & 0.43s & 1.25$\times$ \\
		DeiT-S (50\%) &
		9.81ms & 7.34ms & 1.34$\times$ &
		107.42ms & 81.05ms & 1.33$\times$ &
		1.72s & 1.20s & 1.43$\times$ \\
		DeiT-B (55\%) &
		30.73ms & 21.36ms & 1.44$\times$ &
		378.63ms & 273.94ms & 1.38$\times$ &
		5.71s & 3.81s & 1.50$\times$ \\
		DeiT-B (20\%) &
		30.73ms & 27.72ms & 1.11$\times$ &
		378.63ms & 342.34ms & 1.11$\times$ &
		5.71s & 5.12s & 1.11$\times$ \\
		\midrule
		Swin-T (45\%) &
		14.00ms & 11.66ms & 1.20$\times$ &
		131.78ms & 111.00ms & 1.19$\times$ &
		2.70s & 2.28s & 1.18$\times$ \\
		Swin-S (50\%) &
		24.82ms & 19.60ms & 1.27$\times$ &
		241.27ms & 186.39ms & 1.29$\times$ &
		4.91s & 3.76s & 1.30$\times$ \\
		Swin-B (55\%) &
		36.71ms & 28.14ms & 1.30$\times$ &
		376.82ms & 288.13ms & 1.31$\times$ &
		7.52s & 5.75s & 1.31$\times$ \\
		Swin-B (20\%) &
		36.71ms & 33.37ms & 1.10$\times$ &
		376.82ms & 341.97ms & 1.10$\times$ &
		7.52s & 6.85s & 1.10$\times$ \\
		\midrule
		ConvNeXt-T (45\%) &
		12.07ms  & 9.43ms & 1.28$\times$ &
		133.20ms & 106.44ms & 1.25$\times$ &
		1.93s & 1.48s & 1.30$\times$ \\
		ConvNeXt-S (50\%) &
		21.62ms & 15.20ms & 1.42$\times$ &
		248.75ms & 172.96ms & 1.44$\times$ &
		3.37s & 2.16s & 1.55$\times$ \\
		ConvNeXt-B (55\%) &
		32.39ms & 21.75ms & 1.49$\times$ &
		384.87ms & 249.95ms & 1.54$\times$ &
		5.64s & 3.30s & 1.71$\times$ \\
		ConvNeXt-B (20\%) &
		32.39ms & 28.49ms & 1.14$\times$ &
		384.87ms & 340.90ms & 1.13$\times$ &
		5.64s & 4.91s & 1.15$\times$ \\
		\bottomrule
	\end{tabularx}
\caption{Comparison of runtime performance across different hardware: \textbf{H200 (GPU)}: high-end NVIDIA Tensor Core GPUs; \textbf{T4 (GPU)}: cost-efficient NVIDIA Tesla GPUs for inference; and \textbf{E5 (CPU)}: Intel Xeon E5-2680v4 processors. We report \textbf{Full} and \textbf{VBP} runtimes in milliseconds, along with the resulting \textbf{Speed-up} factors. Our method consistently improves inference latency across diverse hardware environments, reaching up to 1.71$\times$ speed-ups.}

\end{table}

\clearpage

\twocolumn

\section{Activation Distributions Across Layers}
\label{sec:pre-post-act}
\begin{figure}[h]
	\centering

	\begin{subfigure}[t]{\columnwidth} 
		\centering
		\includegraphics[width=\columnwidth]{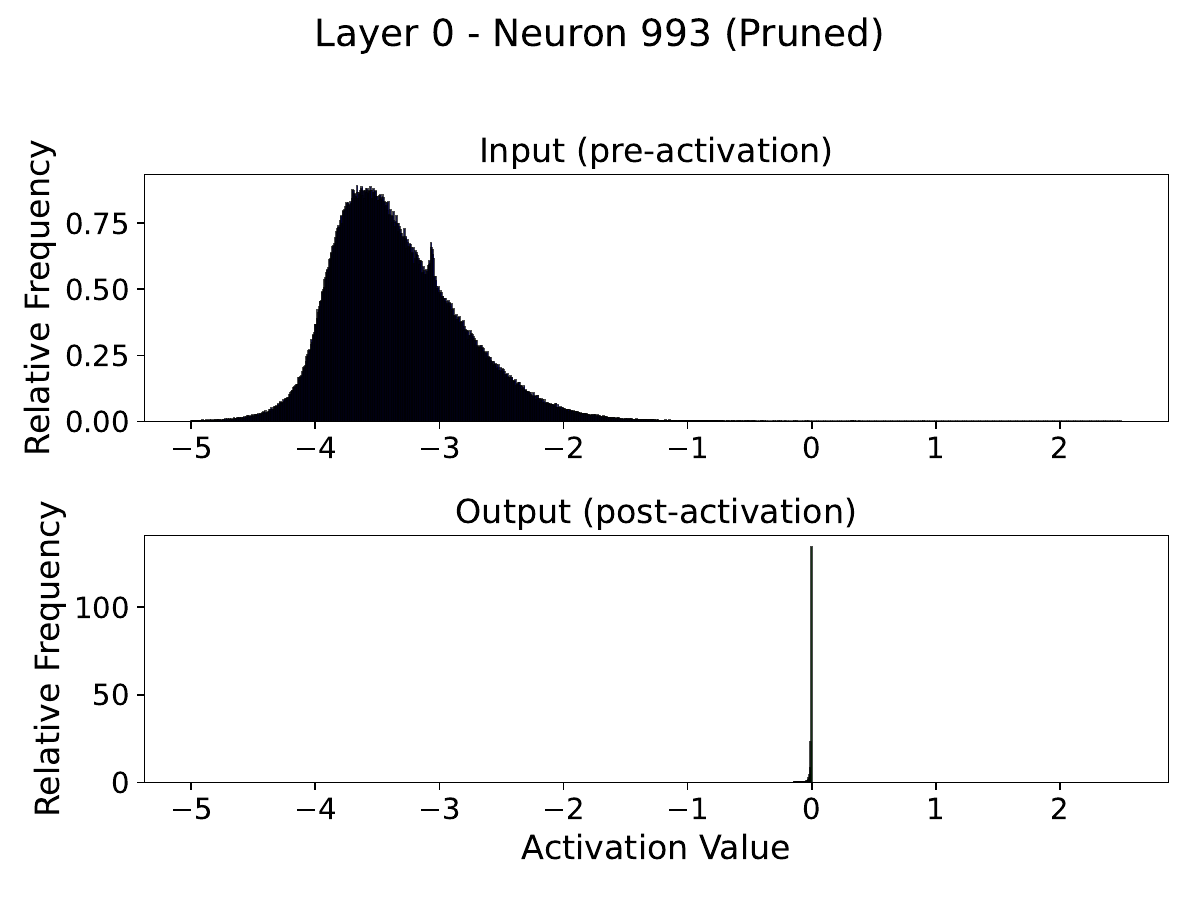}
	\end{subfigure}
	
	\begin{subfigure}[t]{\columnwidth} 
		\centering
		\includegraphics[width=\columnwidth]{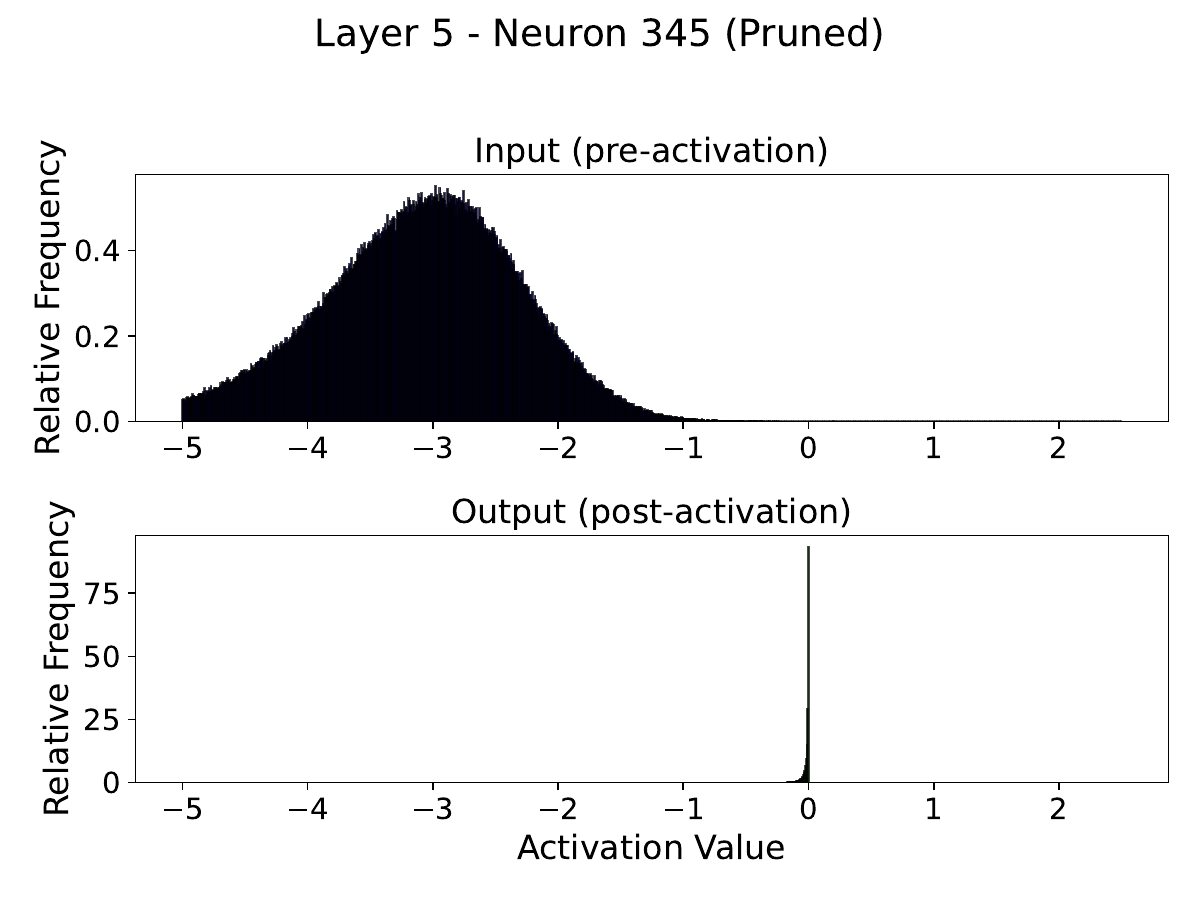}
	\end{subfigure}
	
	\begin{subfigure}[t]{\columnwidth} 
		\centering
		\includegraphics[width=\columnwidth]{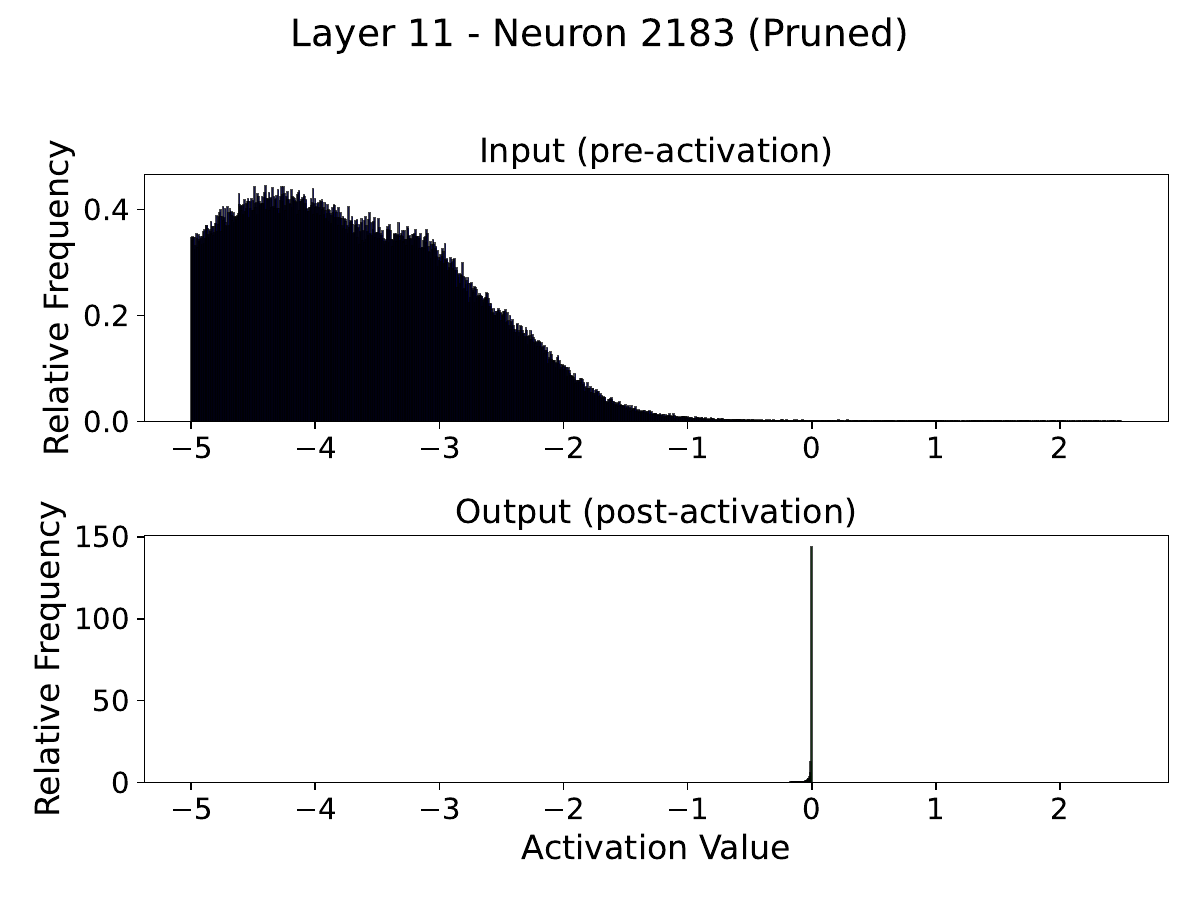}
	\end{subfigure}
	
	\caption{Visualization of activation distributions before and after the non-linearity in pruned neurons throughout different layers.}
	\label{fig:variance-based-pruning}
\end{figure}

\begin{figure}[t]
	\centering

	\begin{subfigure}[t]{\columnwidth} 
		\centering
		\includegraphics[width=\columnwidth]{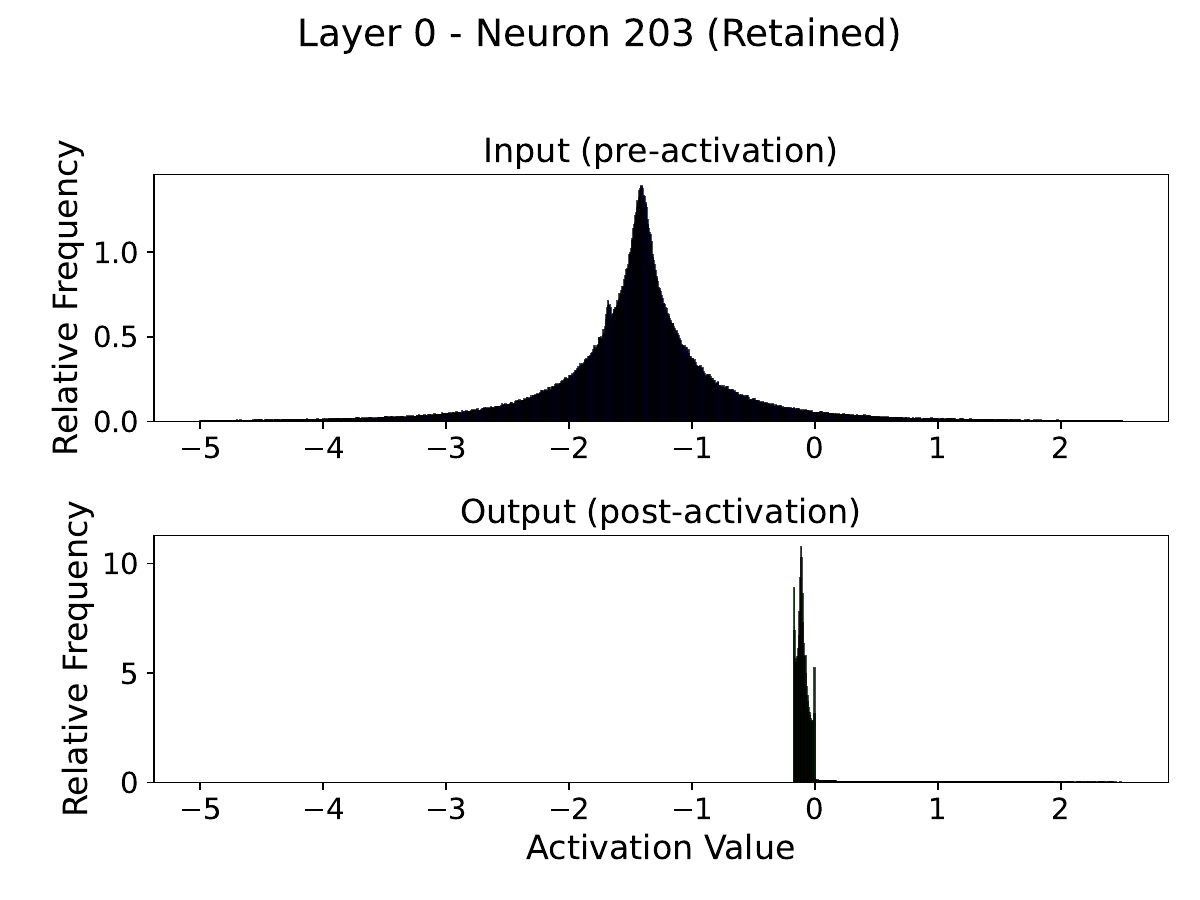}
	\end{subfigure}
	
	\begin{subfigure}[t]{\columnwidth} 
		\centering
		\includegraphics[width=\columnwidth]{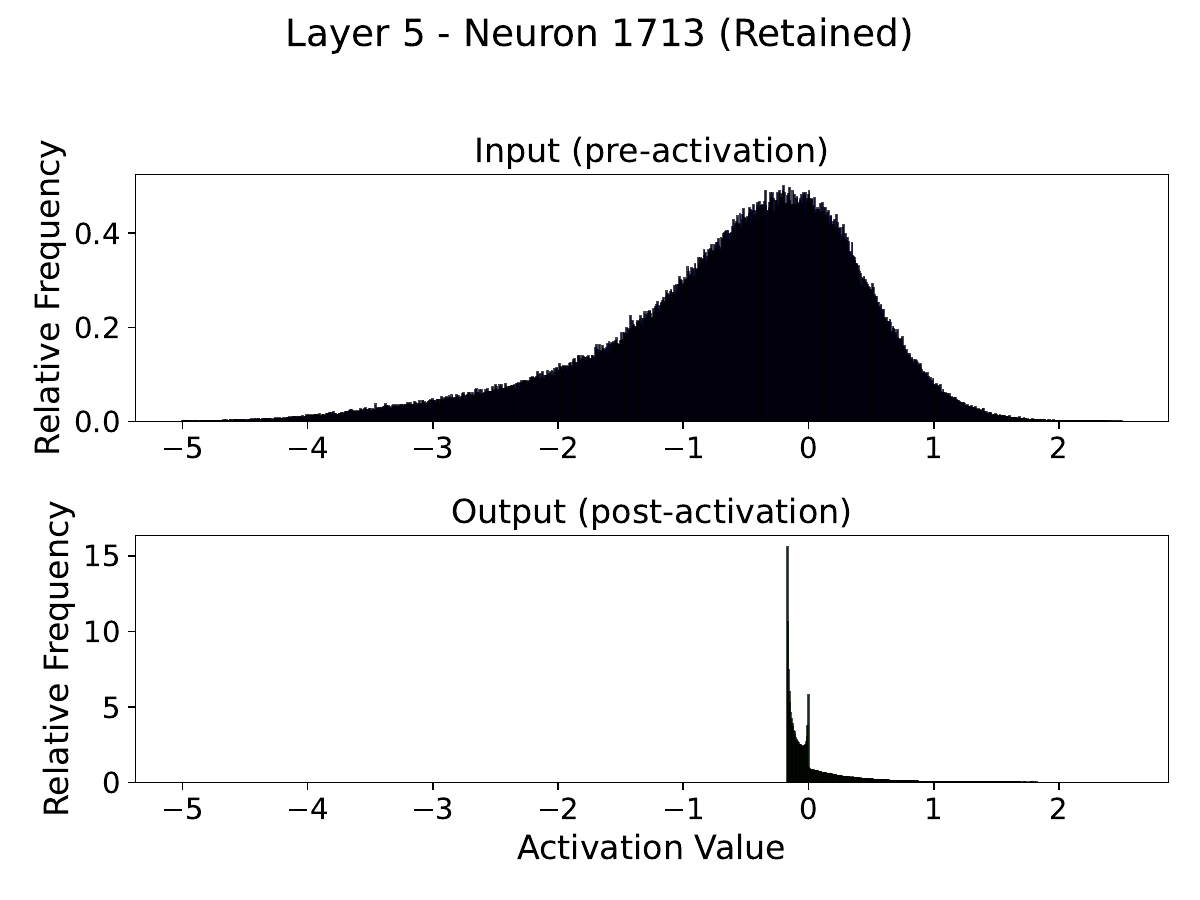}
	\end{subfigure}
	
	\begin{subfigure}[t]{\columnwidth} 
		\centering
		\includegraphics[width=\columnwidth]{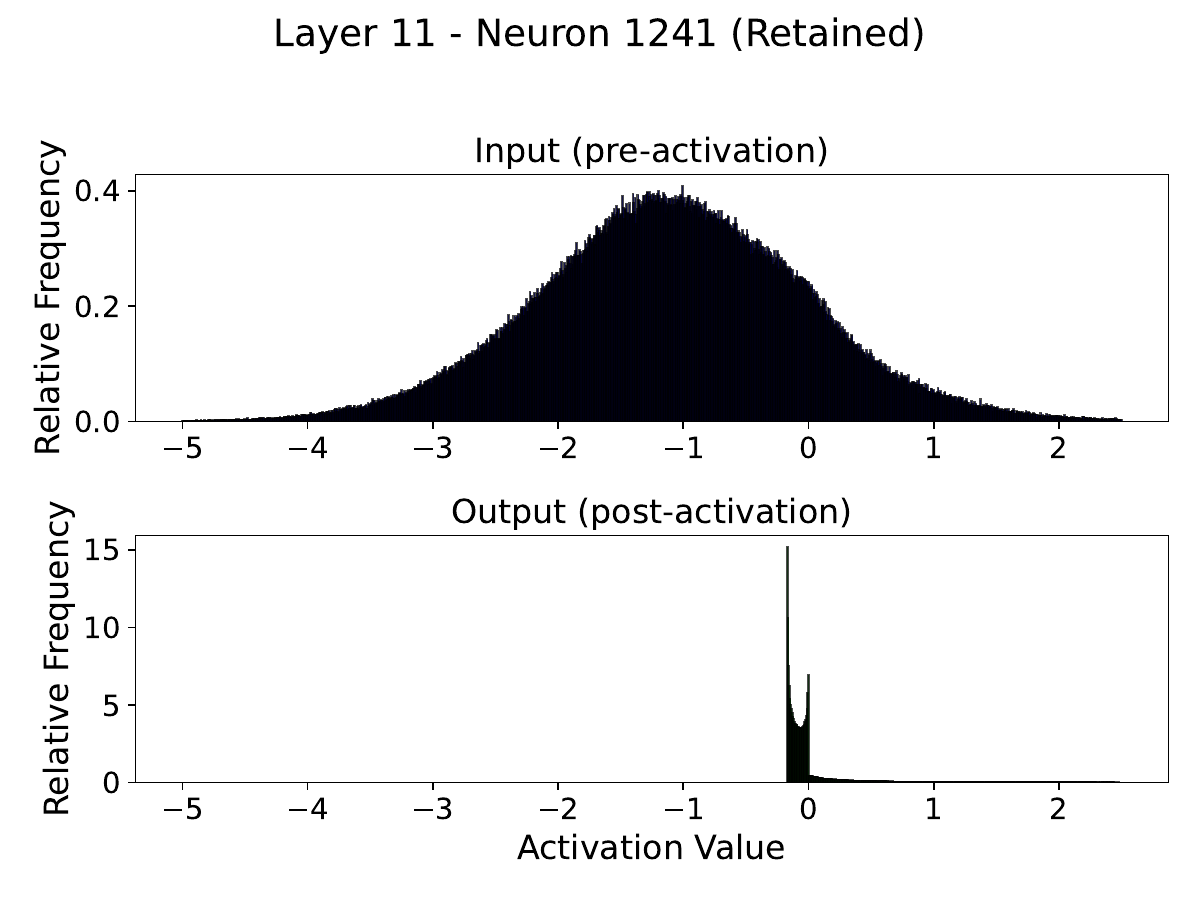}
	\end{subfigure}

	\caption{Visualization of activation distributions before and after the non-linearity in retained neurons throughout different layers.}
	\label{fig:variance-based-pruning}
\end{figure}

\end{document}